\documentclass[lettersize,journal]{IEEEtran}
\usepackage{amsmath,amsfonts}
\usepackage{algorithmic}
\usepackage{algorithm}
\usepackage{array}
\usepackage[caption=false,font=footnotesize,labelfont=sf,textfont=sf]{subfig}
\usepackage{textcomp}
\usepackage{stfloats}
\usepackage{url}
\usepackage{verbatim}
\usepackage{graphicx}
\usepackage{cite}
\usepackage[table,xcdraw]{xcolor}
\usepackage{tabularx}
\usepackage{colortbl}
\usepackage{adjustbox}
\usepackage{multirow} 
\usepackage{booktabs}
\usepackage{caption}
\usepackage{subcaption}
\usepackage{capt-of}
\usepackage[citecolor=blue, colorlinks]{hyperref}
\usepackage{tikz}
\DeclareRobustCommand{\bluedashbox}{\tikz[baseline=0.3ex]{\draw[cyan!60!black, thick, fill=cyan!10] (0,0) rectangle (1.2em, 0.8em);}}
\DeclareRobustCommand{\orangebox}{\tikz[baseline=0.3ex]{\draw[orange!50, thick, fill=orange!10] (0,0) rectangle (1.2em, 0.8em);}}
\DeclareRobustCommand{\greenbox}{\tikz[baseline=0.3ex]{\draw[teal!80!green, thick, fill=green!10] (0,0) rectangle (1.2em, 0.8em);}}

\usepackage{soul}

\begin{document}

\title{SPARE-GS: Structural Parsimony and Resource Efficiency for 3D Gaussian Splatting}

\author{Zhang Chen, Shuai Wan, Fuzheng Yang, Jiazhi Xia, Weiyao Lin,~\IEEEmembership{Senior Member,~IEEE} \\ and Junhui Hou,~\IEEEmembership{Senior Member,~IEEE}
\thanks{Zhang Chen is with the School of Electronics and Information, Northwestern Polytechnical University, Xi'an 710129, China, and also with the Department of Computer Science, City University of Hong Kong, Hong Kong SAR, China. (email: chenzhang@mail.nwpu.edu.cn)}
\thanks{Shuai Wan is with the School of Electronics and Information, Northwestern Polytechnical University, Xi'an 710129, China (e-mail: swan@nwpu.edu.cn).}
\thanks{F. Yang is with the School of Telecommunication Engineering, Xidian University, Xi'an 710071, China (e-mail: fzhyang@mail.xidian.edu.cn).}
\thanks{J. Xia is with the School of Computer Science and Engineering, Central South University, China (email: xiajiazhi@csu.edu.cn)}
\thanks{W. Lin is with the Department of Electrical Engineering, Shanghai Jiao Tong University, Shanghai, China (e-mail: wylin@sjtu.edu.cn).}
\thanks{J. Hou is with the Department of Computer Science, City University of Hong Kong, Hong Kong SAR, China (e-mail: jh.hou@cityu.edu.hk).}
\thanks{This work was supported in part by the TCL Science and Technology Innovation Fund, the NSFC under Grants 62371358 and 62422118, and the Hong Kong Research Grants Council under Grants 11220426, 11219324, and N\_CityU1114/25.}
}

\maketitle
\begin{abstract}
3D Gaussian Splatting (3DGS) achieves high-fidelity novel view synthesis in real-time; however its training efficiency and representation compactness are hindered by excessive primitive proliferation. To address this challenge, we formulate the structural evolution of 3DGS as a global budget-constrained optimization problem and derive an optimality condition, which requires the marginal utility of structural resources to be balanced across spatial regions under a finite primitive budget. Based on this formulation, we propose SPARE-GS, a general \textit{plug-and-play} framework that dynamically aligns the distribution of 3D Gaussian primitives with regional representational demand. SPARE-GS estimates capacity-normalized regional demand, assigns adaptive target quotas, and uses regional budget deviations to coordinate densification, pruning and adaptive termination toward a more balanced structural allocation. Extensive experiments across standard, accelerated, and structure-enhanced 3DGS pipelines demonstrate that SPARE-GS reduces the Gaussian count and training time by an average of 30.38\% and 23.81\%, respectively, while improving the average PSNR. Moreover, the resulting compact representations reduce downstream processing time and improve the rate-distortion performance of diverse compression and pruning methods, demonstrating the broad applicability of global structural budget regulation.
We refer readers to the project page at \url{ https://zhangchen2022.github.io/SPARE-GS.github.io/} for the source code and more visual results.
\end{abstract}

\begin{IEEEkeywords}
3D Gaussian Splatting; Novel View Synthesis; Budgeted Optimization; Structural Parsimony; Efficiency; Compression
\end{IEEEkeywords}

\section{Introduction}
\IEEEPARstart{3}{D} Gaussian Splatting (3DGS) \cite{kerbl20233d} has emerged as an effective approach for novel view synthesis, achieving a favorable balance between photo-realistic quality and real-time rendering. By explicitly representing scenes with anisotropic 3D Gaussian primitives and utilizing a differentiable rasterization pipeline, 3DGS has become a widely adopted method for 3D scene representation~\cite{shen2025gamba,zhou2025gpsgaussianplus,wu2025deferredgs,lei2025gaussnav}. 
However, this explicit representation introduces considerable resource overhead~\cite{ fang2026efficient, wang2026freesplatplus,bai2026plug}. High-fidelity reconstructions typically require massive Gaussian primitives to capture complex geometries and fine view-dependent appearances~\cite{niedermayr2024compressed, du2026mobile}. This dense primitive count increases model size, posing challenges for mobile deployment and large-scale scene modeling, while raising memory consumption and training time~\cite{chen2025haifgs, chen2026feedforward, niemeyer2024radsplat, ren2025octreegs}. 
Furthermore, the conventional fixed-iteration training schedule often continues optimization even after the scene structure and reconstruction loss have converged. 
Consequently, advancing the practicality of 3DGS becomes a problem of comprehensive cost control across training time, computational resources, memory footprint, and storage capacity, a challenge that has driven recent efforts to explore various optimization strategies~\cite{chen2025fast,fang2024mini}.

\begin{figure}[t] 
    \centering
    \begin{minipage}[b]{0.47\textwidth}
        \centering
        \includegraphics[width=\textwidth]{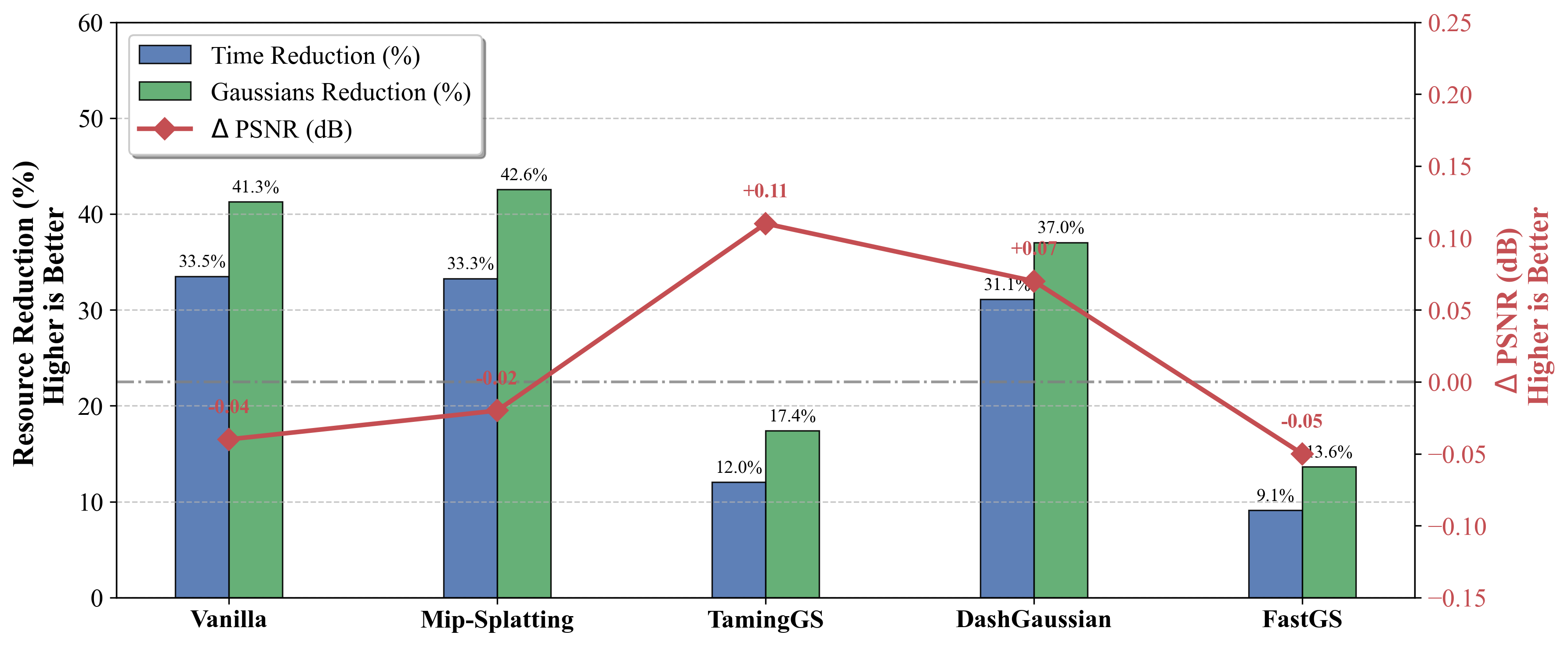} 
        \vspace{0.2cm}
        \centerline{\small (a) Efficiency and Quality Gains during 3DGS Training}
    \end{minipage}
    \\

    \begin{minipage}[b]{0.45\textwidth} 
        \centering
        \includegraphics[width=\textwidth]{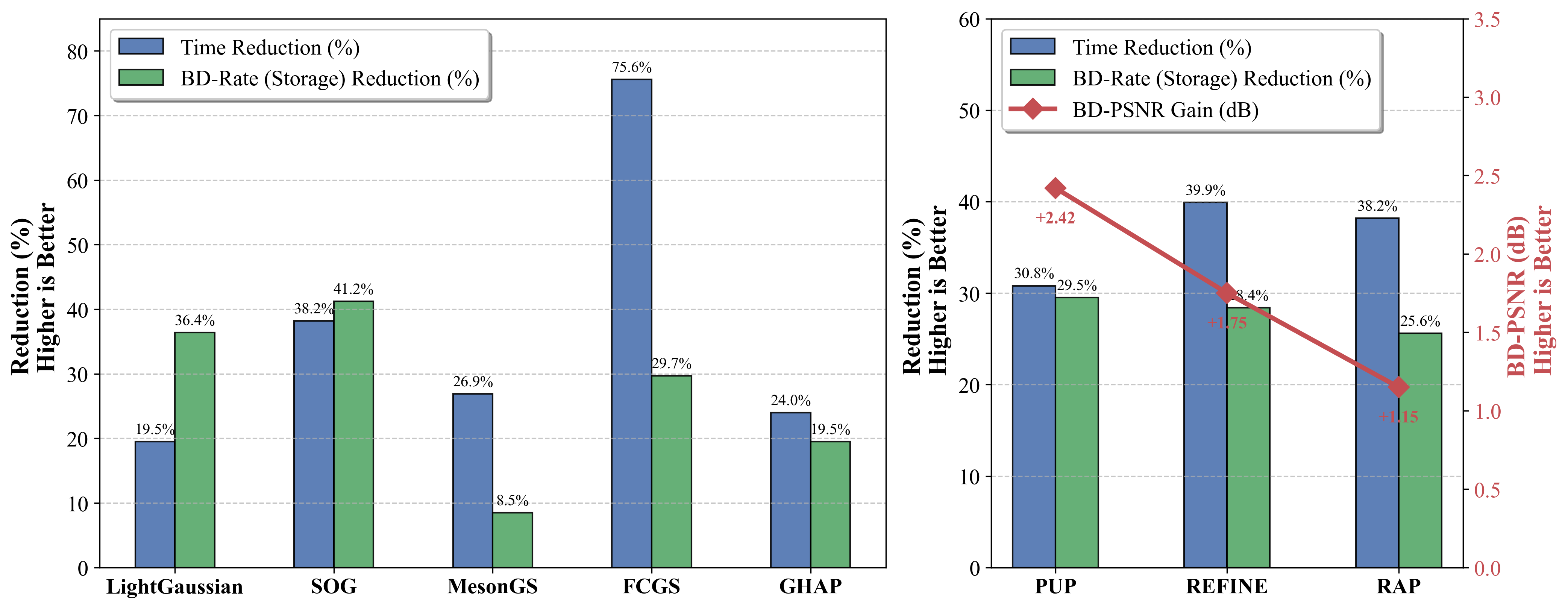}
        \vspace{-0.2cm}
        \centerline{\small (b) Benefits on Post-processing Compression \& Pruning}
    \end{minipage}
    \vspace{0.3cm}

\caption{Performance gains of incorporating our SPARE-GS into diverse off-the-shelf pipelines under various tasks.
(a) Training: across diverse 3DGS pipelines, SPARE-GS reduces training time and Gaussian counts without sacrificing rendering quality. 
(b) Post-Processing: this compact representation benefits downstream tasks, delivering substantial storage and latency savings for various compression and pruning frameworks.}
\label{fig:comprehensive_teaser}
\end{figure}

Existing methods have attempted to improve 3DGS efficiency from two aspects: optimization of structural evolution and representation compression. 
Structural evolution optimized methods improve 3DGS efficiency by reducing optimization cost, refining density control, or redesigning the underlying representation. Existing approaches employ strategies such as resolution scheduling, sparse computation, iteration reduction, multi-view-consistent densification, targeted pruning, and modified rendering formulations~\cite{chen2025dashgaussian,hanson2025speedy,ren2025fastgs,rota2024revising,zhang2024pixel,cheng2024gaussianpro}. Although these approaches improve different aspects of training, most structural decisions still rely on local heuristics and lack explicit spatial budget coordination.

Representation compression primarily targets storage and transmission costs. It can be broadly divided into training-integrated and post-training approaches. Training-integrated methods, such as HAC\cite{chen2024hac}, jointly optimize compact representations, masking or quantization mechanisms, and entropy models during scene reconstruction. Post-training methods instead operate on a pretrained Gaussian representation, including compression frameworks that reorganize, transform, quantize, or entropy-code Gaussian attributes~\cite{morgenstern2024compact,fan2024lightgaussian,xie2024mesongs,wang2025gaussian}, as well as post-hoc pruning methods that remove low-importance primitives and optionally refine the remaining representation~\cite{hanson2025pup,chen2026refine,yang2026rap}. 
These two categories are closely related to the structural organization of the Gaussian representation. Post-training compression is strongly influenced by the redundancy and spatial organization inherited from the pretrained Gaussian representation.

Overall, the organization of Gaussian primitives directly affects the computational cost of optimization and the amount of redundancy inherited by downstream compression and pruning. However, this structure is typically governed by local density-control decisions during training or treated as a fixed input afterward, rather than being explicitly managed as a finite global resource. This motivates a fundamental question: \textit{\textit{how can the structural budget of 3DGS be globally regulated to improve training efficiency and representational compactness?}}

We argue that achieving structural parsimony in 3DGS requires formulating structural optimization through the lens of budget management, rather than relying on unconstrained primitive-level decisions. Each Gaussian primitive represents a unit of computational and memory cost, while different spatial regions exhibit varying representational demands. Regions with complex geometry, high-frequency textures, or strong multi-view visibility naturally justify higher budget utilization; conversely, regions that already contain sufficient primitives offer diminishing quality gains from further densification.
Consequently, conventional gradient-threshold-based densification fails to enforce global budget regulation, which inevitably leads to redundant primitive proliferation in some areas and structural under-representation in others, ultimately inflating the overall cost.

Based on this perspective, we formulate the structural evolution of 3DGS as a global budget-constrained optimization problem and introduce SPARE-GS, a \textit{plug-and-play} framework for structural regulation. SPARE-GS estimates regional representational demand from aggregated multi-view statistics, converts it into capacity-aware target quotas, and uses regional budget deviations to coordinate densification and pruning. Photometric and structural stability further provide a principled signal for adaptive termination.

Extensive experiments across diverse 3DGS training pipelines demonstrate substantial reductions in training time and Gaussian count, while improving rendering fidelity. The resulting compact representations also improve the efficiency and rate-distortion performance of downstream compression and pruning methods, as shown in Fig.~\ref{fig:comprehensive_teaser}.

The main contributions are summarized as follows.

\begin{itemize}
\item We formulate the structural evolution of 3DGS as a global budget-constrained optimization problem and derive a optimality condition, which requires the marginal utility of structural resources to be balanced across spatial regions under a finite primitive budget.

\item Guided by this condition, we propose SPARE-GS, which estimates regional marginal demand, constructs capacity-aware target quotas, and uses regional budget deviations to steer densification, pruning, and adaptive termination toward a more balanced structural allocation.

\item We validate SPARE-GS across diverse 3DGS training pipelines and downstream tasks, demonstrating substantial reductions in training time and Gaussian count while preserving rendering fidelity and improving the efficiency of subsequent post-processing.
\end{itemize}

The rest of this paper is organized as follows. Section \ref{sec2} reviews previous work in 3DGS. Section \ref{sec3} outlines the formulation of the problem and the fundamental principles behind SPARE-GS. The details of SPARE-GS are described in Section \ref{sec4}. Section \ref{sec5} introduces extensive experimental results. Finally, Section \ref{sec6} concludes this paper.\par

\section{Related Work}
\label{sec2}
\subsection{Optimization of 3DGS}
Existing methods improve 3DGS efficiency by jointly reducing optimization cost and refining the evolution of Gaussian representations. Speedy-Splat~\cite{hanson2025speedy} exploits sparse pixels and primitives to avoid unnecessary computation, while DashGaussian~\cite{chen2025dashgaussian} coordinates rendering resolution and primitive growth throughout optimization. FastGS~\cite{ren2025fastgs} introduces multi-view-consistent densification and targeted pruning to suppress irrelevant primitives and reduce training cost. Other studies focus more directly on density control. Revising Densification~\cite{rota2024revising} introduces a pixel-error-driven criterion, Pixel-GS~\cite{zhang2024pixel} incorporates pixel coverage into multi-view gradient aggregation, and AbsGS~\cite{ye2024absgs} alleviates gradient collisions using homodirectional view-space gradients. GaussianPro~\cite{cheng2024gaussianpro} progressively propagates Gaussians using multi-view stereo priors, whereas TamingGS~\cite{mallick2024taming} performs constructive densification under a prescribed global Gaussian budget. MCMC-based 3DGS~\cite{kheradmand20243d} further reformulates Gaussian evolution as a sampling process through transitions and relocalization.

Several methods improve training behavior by redesigning the representation or rendering formulation~\cite{wang2025steepest}. Mip-Splatting~\cite{yu2024mip} introduces 3D smoothing and 2D Mip filtering to reduce aliasing under varying sampling rates. Scaffold-GS~\cite{lu2024scaffoldgs} organizes local Gaussians around neural anchors and predicts view-dependent attributes on demand, while 2DGS~\cite{huang20242d} replaces volumetric Gaussians with oriented Gaussian disks to improve geometric consistency. Although these methods improve training efficiency or rendering quality, explicit coordination of a finite global budget across spatial regions according to estimated demand remains insufficiently explored.

\subsection{Compression of 3D Gaussian Representations}
Representation compression targets the storage, transmission, and deployment costs of 3DGS. It can be broadly divided into training-integrated and post-training approaches. Training-integrated methods jointly optimize the scene representation and its coding model. Scaffold-GS~\cite{lu2024scaffoldgs} provides a structured anchor-based representation, upon which HAC~\cite{chen2024hac} models spatial context through a hash grid for entropy coding. HAC++~\cite{chen2025hacplus} extends this formulation with inter- and intra-anchor context modeling, adaptive quantization, and masking. Related approaches employ structured parameterizations, hierarchical spatial organization, or compact basis functions to reduce representation redundancy~\cite{ren2025octreegs,chen2025megs}.

Post-training compression operates on an already optimized Gaussian representation. Attribute-oriented frameworks reduce storage by transforming, quantizing, or entropy-coding Gaussian parameters. Compact-GS~\cite{lee2024compact} and CompGS~\cite{liu2024compgs} compress Gaussian attributes through compact parameterization and quantization, while SOG~\cite{morgenstern2024compact} maps Gaussian attributes onto structured grids for video-based coding. LightGaussian~\cite{fan2024lightgaussian} combines importance-based pruning, recovery, spherical-harmonic distillation, and vector quantization. MesonGS~\cite{xie2024mesongs}, FCGS~\cite{chen2025fast}, and GHAP~\cite{wang2025gaussian} further improve rate-distortion performance through attribute transformation, structured coding, or geometry-aware processing.
A complementary group of post-hoc pruning methods focuses on reducing the number of primitives before or alongside subsequent coding. PUP~\cite{hanson2025pup} estimates the reconstruction sensitivity of Gaussian spatial parameters and applies iterative prune-refine operations. REFINE~\cite{chen2026refine} employs an efficient rendering-aware importance approximation, while RAP~\cite{yang2026rap} removes redundant primitives according to its pruning criterion. Other methods use opacity, scale, gradients, or rendering contribution to rank and eliminate low-importance Gaussians~\cite{ali2024trimming,ali2025elmgs,girish2024eagles}. The effectiveness of post-training compression and pruning is influenced by the redundancy and spatial organization inherited from the pretrained representation. 

Overall,  the Gaussian structure affects the computational cost during optimization and determines the redundancy presented to downstream compression and pruning. However, this structure is commonly shaped by local density-control decisions or treated as a fixed input after training, rather than being explicitly managed as a finite global resource.

\section{Problem Formulation} \label{sec3}

This section establishes the theoretical foundation of SPARE-GS by framing 3DGS structural evolution as a constrained optimization problem. We first formulate the densification and pruning process as a budget-constrained mixed-integer nonlinear programming (MINLP) task, uncovering the limitations of conventional heuristic-based training (Section \ref{sec3.1}). To address these inefficiencies, we relax this formulation into a regional budget optimization framework governed by the principle of diminishing marginal utility (Section \ref{sec3.2}). By deriving the optimal structural state through Karush-Kuhn-Tucker (KKT) conditions~\cite{gordon2012karush}, we prove that structural parsimony is achieved when the marginal rendering utility is balanced across different spatial regions (Section \ref{sec3.3}). These insights provide the guiding principle for the practical resource-regulation mechanism detailed in Section \ref{sec4}.

\subsection{Rethinking of 3DGS Training}
\label{sec3.1}
A 3D scene in 3DGS is represented by a set of $N$ Gaussian primitives ${\cal G} = \left\{ {{G_i} = ({\boldsymbol{\mu} _i}, {\boldsymbol{s}_i}, {\boldsymbol{q}_i},{\boldsymbol{c}_i},{\alpha _i})} \right\}_{i = 1}^N$.
Each primitive is defined by its center ${\boldsymbol{\mu} _i} \in \mathbb{R}^3$, spatial scaling ${\boldsymbol{s} _i} \in \mathbb{R}^{3}$, rotation ${\boldsymbol{q} _i} \in \mathbb{R}^4$, 
color spherical harmonics coefficients ${\boldsymbol{c} _i} \in \mathbb{R}^{3 \times 16}$, and opacity $\alpha_i \in \mathbb{R}$~\cite{kerbl20233d}. During rendering, these primitives are projected and depth-sorted to perform alpha-blending along each pixel ray $p$, computing the final color $\boldsymbol{C}(p)$ via the accumulated transmittance $T_i(p)$ \cite{kopanas2021point}, formulated as
\begin{equation}
\boldsymbol{C}(p) = \sum_{i \in \mathcal{N}} \tilde{\boldsymbol{c}}_i \tilde\alpha_i(p) T_i(p),\quad  T_i(p) = \prod_{j=1}^{i-1} (1 - \tilde\alpha_j(p)),
  \label{eq:2}
\end{equation}
where $\mathcal{N}$ is the depth-sorted set of primitives overlapping pixel $p$, $\tilde{\boldsymbol{c}_i}$ is the color computed via spherical harmonics, 
and $\tilde{\alpha}_i(p)$ is the Gaussian footprint evaluated in pixel $p$ multiplied by the standalone opacity $\alpha_i$.

During training, 3DGS optimizes these parameters by minimizing the photometric loss between the synthesized and ground-truth images over $\mathcal{V}$ viewpoints:

\begin{equation}
\mathcal{L}(\mathcal{G})
=
\frac{1}{|\mathcal{V}|}
\sum_{v \in \mathcal{V}}
L\left(
\mathcal{R}_v(\mathcal{G}), \mathcal{I}_v
\right),
\label{eq:loss}
\end{equation}
where $\mathcal{R}_v(\cdot)$ denotes differentiable Gaussian rasterization at view $v$, $\mathcal{I}_v$ is the corresponding ground-truth image, and $L(\cdot)$ denotes the photometric loss.

Minimizing Eq.~(\ref{eq:loss}) is not a standard continuous optimization problem; it couples parameter tuning with discrete structural operations, such as primitive cloning, splitting, and pruning. To describe this structural decision process, we consider a structural update at training step $t$. Let $\mathcal{A}^{(t)}$ denote the set of Gaussian primitives that are active before the update, and  $\mathcal{C}^{(t)}$ the set of new candidate primitives proposed by densification. The complete set of admissible primitives for the structural decision is formulated as:
\begin{equation}
\bar{\mathcal{G}}^{(t)} = \mathcal{A}^{(t)} \cup \mathcal{C}^{(t)} = \left\{ G_i^{(t)} \right\}_{i=1}^{{\bar{N}^{(t)}}},
\label{eq:candidate_set}
\end{equation}
where ${\bar{N}^{(t)}}$ is the total number of admissible primitives.

For each primitive $G_i^{(t)} \in \bar{\mathcal{G}}^{(t)}$, we introduce a binary structural variable $\sigma_{i}^{(t)} \in \{0,1\}$. If $\sigma_{i}^{(t)} = 1$, the primitive is selected to remain active after the update, participating in subsequent rendering and optimization. Conversely, if $\sigma_{i}^{(t)}=0$, it is deactivated and excluded from training. The updated active Gaussian set is calculated by:
\begin{equation}
\mathcal{\tilde{G}}^{(t)}(\boldsymbol{\sigma}^{(t)}) = \left\{ G_i^{(t)} \in \bar{\mathcal{G}}^{(t)} \mid \sigma_{i}^{(t)}=1 \right\},
\label{eq:active_set}
\end{equation}
where $\boldsymbol{\sigma}^{(t)} = \{\sigma_{i}^{(t)}\}_{i=1}^{{\bar{N}^{(t)}}}$ denotes the structural activation states. The assignment of $\boldsymbol{\sigma}^{(t)}$ governs the entire structural evolution: it determines whether existing primitives in $\mathcal{A}^{(t)}$ are retained or pruned, and whether new candidates in $\mathcal{C}^{(t)}$ are accepted or discarded. Consequently, densification and pruning are components of a unified selection process over the admissible set $\bar{\mathcal{G}}^{(t)}$.

By incorporating structural decisions into the photometric optimization, we reformulate 3DGS training as a resource-constrained MINLP task:
%\begin{equation}
\begin{align}
\label{eq:minlp}
\min_{\bar{\mathcal{G}}^{(t)}, \boldsymbol{\sigma}^{(t)}} \quad & \mathcal{L}\left(\mathcal{\tilde{G}}^{(t)}\left(\boldsymbol{\sigma}^{(t)}\right)\right), \\
\text{s.t.} \quad & \sum_{i=1}^{{\bar{N}^{(t)}}} \sigma_{i}^{(t)} \leq B^{(t)}, \quad \sigma_{i}^{(t)} \in \{0, 1\}, \nonumber
\end{align}
%\end{equation}
where $B^{(t)}$ is the maximum number of active primitives allowed in step $t$ in the scene, and $\sum_{i=1}^{{\bar{N}^{(t)}}} \sigma_{i}^{(t)} \leq B^{(t)}$ limits the structural complexity of the scene representation, which dictates memory consumption, rendering time, training overhead, or storage size.

Compared with Eq.~(\ref{eq:loss}), Eq.~(\ref{eq:minlp}) models Gaussian primitives as finite resources bounded by $B^{(t)}$. By omitting this global constraint $\sum_{i=1}^{{\bar{N}^{(t)}}} \sigma_{i}^{(t)} \leq B^{(t)}$, the original 3DGS pipeline relies on isolated local rules, such as fixed gradient thresholds. Because these unconstrained heuristics operate without global awareness, they fail to balance structural growth across the scene, inevitably causing severe redundancy in high-frequency areas and under-representation elsewhere. Therefore, effective 3DGS optimization goes beyond merely reducing local loss; \textbf{\textit{it involves regulating structural evolution under a finite budget to achieve high representational efficiency.}}

%%%%%%%%%%%%%%%%%%%%%%%%%%%%%%%%%%%%%%%%%%%%%%%%%%%%%%%%%%%%%%%%%%%%%
\subsection{Modeling via Marginal Utility}
\label{sec3.2}
While Eq.~(\ref{eq:minlp}) provides a formal objective for structural evaluation, solving this MINLP problem is computationally intractable during training. Therefore, we introduce a spatial relaxation: instead of formulating binary decisions for millions of primitives, we aggregate them into local spatial regions and optimize the continuous representational budget governed within each region.

Specifically, we subdivide the 3D scene into $K$ disjoint spatial regions $\{\Omega_r\}_{r=1}^{K}$. Let $n_r^{(t)} \geq 0$ represent the number of active primitives budgeted for region $\Omega_r$ at training step $t$, and we approximate the overall optimization objective as the sum of independent regional contributions. Let $U_r\left(n_r^{(t)}\right)$ denote the expected local loss reduction (i.e., utility) achieved by dedicating a structural budget of $n_r^{(t)}$ primitives to region $\Omega_r$. We simply Eq.~(\ref{eq:minlp}) into a region-level budget optimization problem:
\begin{equation}
\begin{aligned}
\max_{\{n_r^{(t)}\}_{r=1}^{K}} \quad & \sum_{r=1}^{K} U_r(n_r^{(t)}), \\
\mathrm{s.t.} \quad & \sum_{r=1}^{K} n_r^{(t)} \leq B^{(t)},~ n_r^{(t)} \geq 0.
\end{aligned}
\label{eq:allocation}
\end{equation}

During training, as more Gaussian primitives are added to $\Omega_r$, the ability to capture high-frequency details eventually saturates. As a result, the additional loss reduction provided by each added primitive decreases. This property corresponds to the law of diminishing marginal utility~\cite{ormazabal1995law}, expressed as
\begin{equation}
\frac{\partial U_r(n_r^{(t)})}{\partial n_r^{(t)}} \geq 0, \quad \frac{\partial^2 U_r(n_r^{(t)})}{\partial {\left(n_r^{(t)}\right)}^2} \leq 0.
\label{eq:marginal}
\end{equation}

Given this diminishing nature, strategically distributing the finite structural budget becomes essential for achieving high representational efficiency. This objective motivates our formal derivation of the optimal representational state in the subsequent section.

%%%%%%%%%%%%%%%%%%%%%%%%%%%%%%%%%%%%%%%%%%%%%%%%%%%%%%%%%%%%%%%%%%%%%%%%

\subsection{Optimal Regulation by KKT Conditions}
\label{sec3.3}
To determine the optimal regional budget configuration $(\boldsymbol{n}^*)^{(t)} = \{(n_1^*)^{(t)}, \dots, (n_K^*)^{(t)}\}$ at training step $t$, we construct the associated Lagrangian function:
\begin{equation}
\begin{split}
\begin{array}{l}
{\cal J}({{\boldsymbol{n}}^{(t)}},{\lambda ^{(t)}},{{\boldsymbol{\nu }}^{(t)}})\\
 = \sum\limits_{r = 1}^K {{U_r}} (n_r^{(t)}) + {\lambda ^{(t)}}({B^{(t)}} - \sum\limits_{r = 1}^K {n_r^{(t)}} ) + \sum\limits_{r = 1}^K {\nu _r^{(t)}} n_r^{(t)},
\end{array}
\end{split}
\label{eq:lagrangian}
\end{equation}
where $\lambda^{(t)}$ denotes the Lagrange multiplier, and $\boldsymbol{\nu}^{(t)} = \{\nu_r^{(t)}\}_{r=1}^K$ are the multipliers of the regional budgets.
According to nonlinear programming theory~\cite{nocedal2006numerical}, if $(\boldsymbol{n}^*)^{(t)}$ represents the optimal budget configuration, it satisfies the KKT conditions:
\begin{itemize}
    \item Stationarity:
\begin{equation}
\frac{\partial U_r((n_r^*)^{(t)})}{\partial n_r^{(t)}} - \lambda^{(t)} + \nu_r^{(t)} = 0,
\label{eq:kkt_stat}
\end{equation}

\item Primal Feasibility:
\begin{equation}
\sum_{r=1}^{K} (n_r^*)^{(t)} \leq B^{(t)}, \quad (n_r^*)^{(t)} \geq 0, 
\label{eq:kkt_primal}
\end{equation}

\item Complementary Slackness:
\begin{equation}
\lambda^{(t)} \left( B^{(t)} - \sum_{r=1}^{K} (n_r^*)^{(t)} \right) = 0, \quad \nu_r^{(t)} (n_r^*)^{(t)} = 0. \label{eq:kkt_comp}
\end{equation}
\end{itemize}

During training, the budget constraint is binding, meaning $\lambda^{(t)} > 0$ and $\sum_{r=1}^{K} (n_r^*)^{(t)} = B^{(t)}$. Furthermore, for any spatial region $\Omega_r$ that actively participates in representing the scene, its optimal budget is strictly positive ($(n_r^*)^{(t)} > 0$). According to the complementary slackness condition, this implies $\nu_r^{(t)} = 0$. Consequently, we derive the optimality criterion:
\begin{equation}
\frac{\partial U_r((n_r^*)^{(t)})}{\partial n_r^{(t)}} = \lambda^{(t)}, \quad \forall r \text{ where } (n_r^*)^{(t)} > 0.
\label{eq:equimarginal}
\end{equation}

Eq.~(\ref{eq:equimarginal}) establishes that in the optimal state, the marginal utility yielded by a unit increase in the structural budget is uniformly equalized across all active regions, converging to the constant $\lambda^{(t)}$. The KKT condition provides a target equilibrium criterion for regional resource allocation. During the intermediate training phase at step $t$, the current structural allocation $n_r^{(t)}$ is typically unbalanced. Their marginal utilities initially exhibit a skewed distribution:
\begin{equation}
\frac{\partial U_1(n_1^{(t)})}{\partial n_1^{(t)}} \neq \frac{\partial U_2({n_2^{(t)}})}{\partial n_2^{(t)}} \neq \cdots \neq \frac{\partial U_K({n_K^{(t)}})}{\partial n_K^{(t)}}.
\label{eq:imbalance}
\end{equation}

While unconstrained heuristics may inadvertently perpetuate this imbalance, the KKT optimality criterion dictates that progressively eliminating this gap is essential for an efficient structural evolution.

Therefore, although analytically computing the exact $\lambda^{(t)}$ is intractable, this optimality condition provides a directional guide. The proposed method in Sec.~\ref{sec4} leverages this principle as an online feedback mechanism. By measuring the relative structural demands, we modulate the densification and pruning operations to actively steer the marginal utilities across regions toward the KKT equilibrium during training:
\begin{equation}
\frac{\partial U_1(n_1^{(t+\delta t)})}{\partial n_1^{(t+\delta t)}} \approx \cdots \approx \frac{\partial U_K({n_K^{(t+\delta t)}})}{\partial n_K^{(t+\delta t)}} \longrightarrow \lambda^{(t+\delta t)},
\label{eq:balance_target}
\end{equation}
where $\delta t > 0$ represents the subsequent training steps. This targeted intervention smoothly steers the system toward the KKT equilibrium without requiring exact analytical solutions.

%%%%%%%%%%%%%%%%%%%%%%%%%%%%%%%%%%%%%%%%%%%%%%%%%%%%%%%%%%%%%%%%%%%%%%%

\begin{figure*}[t]
  \centering
  \includegraphics[width=\linewidth]{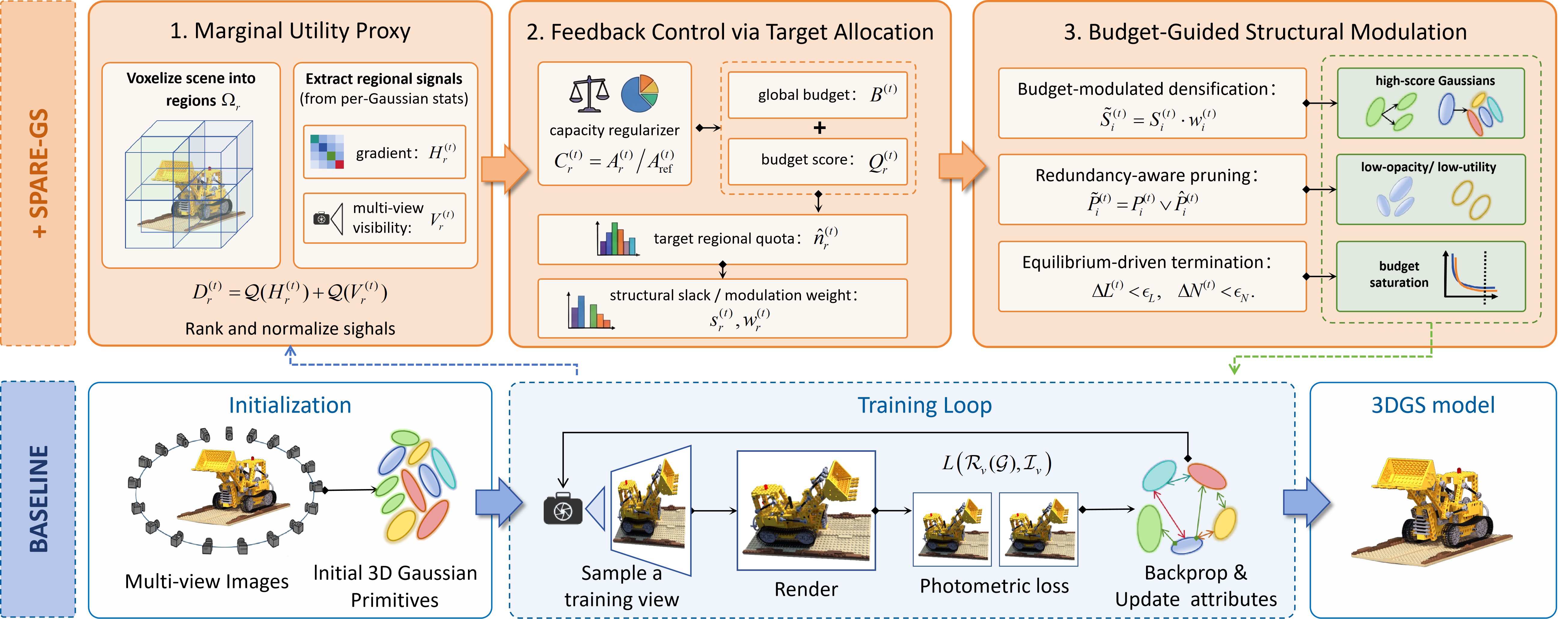}
\caption{\textbf{Overview of the proposed SPARE-GS framework.} To achieve global budgeted-constrained optimization, our method first establishes a \textit{Marginal Utility Proxy} by voxelizing the scene and aggregating regional statistics. Then, we utilizes \textit{Feedback Control via Target Allocation} to dynamically compute capacity-regularized regional quotas. Finally, we apply the \textit{Budget-Guided Structural Modulation} mechanism to guide densification, redundancy pruning, and equilibrium-driven termination. In the diagram, the standard 3DGS baseline pipeline (\bluedashbox) is controlled by our proposed modules (\orangebox), and the specific structural evaluation is guided by our mechanism (\greenbox). }
\label{fig:overview}
\end{figure*}

\section{Proposed method} \label{sec4}
Based on the analysis in Sec.~\ref{sec3}, an efficient structural state is achieved by balancing marginal utility across different regions. To operationalize this, we first detail how to approximate the regional marginal utility by analyzing demand in Sec.~\ref{sec4.1}. Then, we formulate the capacity-aware target budget in Sec.~\ref{sec4.2}, and finally apply these targets to modulate the structural heuristics in Sec.~\ref{sec4.3}. Fig.~\ref{fig:overview} illustrates the overall flowchart of our method.
%%%%%%%%%%%%%%%%%%%%%%%%%%%%%%%%%%%%%%%%%%%%%%%%%%%%%%%%%%%%%%%%%%%%%%%%%%%%%%%%%%%%%%%%%%%%%%%%%

\subsection{Marginal Utility Proxy}
\label{sec4.1}
To allocate resources effectively, we first identify which spatial regions lack structural details. We formulate a proxy to quantify marginal utility using native training statistics.
Generally, the marginal utility of a spatial region reflects its local representational demand, as indicated by Eq.~(\ref{eq:marginal}). A region exhibiting higher unfulfilled demand implies a greater potential for error reduction, thereby translating to a higher marginal utility when additional primitives are allocated. Therefore, we establish a proxy for the regional marginal utility using observable demand statistics natively tracked during the training process.

To evaluate these statistics, we uniformly partition the bounding box of active Gaussians into a $32 \times 32 \times 32$ voxel grid with a small padding. Each voxel defines a local region $\Omega_r$ containing $n_r^{(t)}$ primitives. The regional demand is then computed exclusively across the $K$ non-empty regions.

For each valid region $\Omega_r$, the proxy for marginal utility is formulated as a composite of training statistics:
\begin{equation}
\frac{\partial U_r(n_r^{(t)})}{\partial n_r^{(t)}} \approx D_r^{(t)} = \mathcal{Q}(H_r^{(t)}) + \mathcal{Q}(V_r^{(t)}),
\label{eq:demand}
\end{equation}
where $H_r^{(t)}$ denotes the regional gradient magnitude, indicating unresolved errors, and $V_r^{(t)}$ captures multi-view visibility, indicating global influence of $\Omega_r$. $\mathcal{Q}(\cdot)$ is a rank-normalization function to map these extracted statistics into the interval $[0,1]$, which is defined as $\mathcal{Q}(m_r^{(t)}) = \frac{\mathrm{rank}(m_r^{(t)})}{K}$
with $\mathrm{rank}(m_r^{(t)})$ computing  the ascending rank of an input metric $m_r^{(t)}$ among all $K$ valid regions. In the following, we detail the explicit derivation of each statistical component.

\vspace{0.5em}
\noindent\textbf{Gradient Magnitude.}
A large gradient indicates that the current primitives fail to accurately fit the local image, revealing a strong demand for additional structural details. 
To quantify this, we track the accumulated view-space positional gradient $a_i^{(t)}$ and accumulated visibility count $d_i^{(t)}$ during differentiable rasterization. 
Since large primitives accumulate higher gradients due to their broad screen coverage, we compute a footprint-normalized gradient to avoid size-induced bias. 
Averaging these normalized scores yields the regional gradient magnitude $H_r^{(t)}$, which evaluates the local convergence state of region $\Omega_r$:
\begin{equation}
H_r^{(t)} = \frac{1}{n_r^{(t)}} \sum_{i=1}^{N^{(t)}} \frac{a_i^{(t)}}{d_i^{(t)}} \mathbb{I}[k_i^{(t)}=r],
\label{eq:gradient_demand}
\end{equation}
where $N^{(t)}$ denotes the total number of active primitives at optimization step $t$, $k_i^{(t)}$ represents the assigned voxel index of the $i$-th primitive based on its spatial coordinate, and $\mathbb{I}[\cdot]$ is the indicator function yielding $1$ if the primitive resides in region $\Omega_r$ (i.e., $k_i^{(t)}=r$) and $0$ otherwise.

\vspace{0.5em}
\noindent\textbf{Multi-view Visibility.}
High multi-view visibility implies that a region strongly influences the rendering across multiple viewpoints, generally requiring greater structural detail to capture its complex appearance. Using the accumulated visibility count $d_i^{(t)}$ tracked during the forward pass for visibility, the regional multi-view visibility $V_r^{(t)}$ is computed as the regional average:
\begin{equation}
V_r^{(t)} = \frac{1}{n_r^{(t)}} \sum_{i=1}^{N^{(t)}} d_i^{(t)} \mathbb{I}[k_i^{(t)}=r].
\label{eq:visibility}
\end{equation}

\subsection{Feedback Control via Target Allocation}
\label{sec4.2}

\vspace{0.5em}
\noindent\textbf{Target Quota Allocation.}
To regulate structural evolution globally, we dynamically scale the global budget $B^{(t)}$ according to the current number of active primitives $N^{(t)}$, without disrupting the native optimization trajectory of the base 3DGS pipeline.

Additionally, the representational capacity of Gaussians varies across regions: for the same photometric demand, regions containing smaller primitives generally require a larger numerical quota. We therefore introduce a geometric capacity regularizer $C_r^{(t)}$ to calibrate the regional demand. For region $\Omega_r$, its average projected footprint is defined as

\begin{equation}
A_r^{(t)} =
\frac{1}{n_r^{(t)}}
\sum_{i=1}^{N^{(t)}}
\left(\rho_i^{(t)}\right)^2
\mathbb{I}\left[k_i^{(t)}=r\right],
\label{eq:region_area}
\end{equation}

where $\rho_i^{(t)}$ denotes the maximum projected 2D radius of primitive $G_i^{(t)}$. We then calculate the geometric capacity regularizer as $C_r^{(t)} = A_r^{(t)}/A^{(t)}_{\mathrm{ref}}$, where $A^{(t)}_{\mathrm{ref}}$ denotes the median projected area across all active regions at step $t$. Accordingly, the capacity-adjusted budget score for region $\Omega_r$ is formulated as $Q_r^{(t)} = D_r^{(t)}/C_r^{(t)}$.

According to our KKT optimality analysis in Eq.~(\ref{eq:equimarginal}), structural resources should be routed to regions with the high unfulfilled demand. Thus, we distribute the global budget $B^{(t)}$ based on the capacity-adjusted demand $Q_r^{(t)}$. By normalizing $Q_r^{(t)}$, we calculate the target distribution ratio $l_r^{(t)}$:
\begin{equation}
l_r^{(t)} = \frac{Q_r^{(t)}}{ \sum_{j=1}^{K} Q_j^{(t)}}.
\label{eq:budget_prob}
\end{equation}
Consequently, the target primitive quota for region $\Omega_r$ is determined by:
\begin{equation}
\hat{n}_r^{(t)} = B^{(t)} l_r^{(t)}.
\label{eq:quota}
\end{equation}
This formulation provides a global reference allocation that guides, rather than strictly enforces, structural growth and effective direction for resource allocation in 3DGS training.

\vspace{0.5em}
\noindent\textbf{Feedback-driven Modulation.}
With the target quota $\hat{n}_r^{(t)}$ established in Eq. (\ref{eq:quota}), we formulate a feedback control signal to modulate the structural evolution of 3DGS. Specifically, we quantify the relative structural deviation, for the region residing in the $i$-th primitive:
\begin{equation}
s^{(t)}_{k_i^{(t)}} = \frac{\hat{n}^{(t)}_{k_i^{(t)}} - n^{(t)}_{k_i^{(t)}}}{\hat{n}_{k_i^{(t)}}+ 1}.
\label{eq:slack}
\end{equation}
A positive slack ($s^{(t)}_{k^{(t)}_i} > 0$) indicates that the region is currently under-budgeted, while a negative slack indicates that the current regional occupancy exceeds its target quota.

To translate the structural deviation into a stable control signal without imposing aggressive discontinuities, the final budget-aware modulation weight $w_i^{(t)}$ assigned to the $i$-th primitive is formulated by coupling the regional slack with its representational demand:
\begin{equation}
w_i^{(t)} =  \Phi\left( s^{(t)}_{k_i^{(t)}} , \tilde{l}_r^{(t)}\right),
\label{eq:budget_weight}
\end{equation}
where $\Phi(\cdot)$ is a bounded sigmoid function that maps the structural deviation into a smooth gating signal, and $\tilde{l}_r^{(t)}$ denotes the target distribution ratio $l_r^{(t)}$ normalized against the maximum value across all valid regions.

%%%%%%%%%%%%%%%%%%%%%%%%%%%%%%%%%%%%%%%%%%%%%%%%%%%%%%%%%%%%%%%%%%%%%%%%%%%%%%%%%%%%%%%%%%
\subsection{Budget-guided Structural Modulation}
\label{sec4.3}
Finally, we apply the derived feedback weights in Eq. (\ref{eq:budget_weight}) to control the actual 3DGS optimization. They softly modulate densification, guide redundancy-aware pruning, and enable adaptive termination when equilibrium is reached.

\vspace{0.5em}
\noindent\textbf{Budget-modulated Densification.}
We incorporate the budget constraint by modulating the base densification score $S_i^{(t)}$ with the effective control weight $w_i^{(t)}$. Specifically, $S_i^{(t)}$ denotes the native densification priority assigned to Gaussian or candidate $i$ by the underlying pipeline at iteration $t$, based on its original method-specific criterion. The resulting budget-aware score is computed as:
\begin{equation}
\tilde S_i^{(t)} = S_i^{(t)} \cdot {w}_i^{(t)}.
\label{eq:ours_score}
\end{equation}
In implementation, the native densification threshold is kept unchanged, and this effective weight softly modulates the primitive-level gradient or importance score. Effectively, this modulated score accelerates densification in under-budgeted regions (${w}_i^{(t)} > 1$) and suppresses it in over-budgeted ones (${w}_i^{(t)} < 1$), continuously steering the structural growth toward the target allocation.

\vspace{0.5em}
\noindent\textbf{Redundancy-aware Pruning.}
To prevent the progressive accumulation of structural redundancy in over-budgeted and low-response regions, we formulate a conservative pruning constraint $\hat P_i^{(t)}$ for each primitive $G_i^{(t)}$:
\begin{equation}
\hat P_i^{(t)} = \mathbb{I}[n^{(t)}_{k_i^{(t)}} > \hat{n}^{(t)}_{k^{(t)}_i}] \cdot \mathbb{I}[\alpha^{(t)}_i < \tau_{\alpha}] \cdot \mathbb{I}[\frac{a_i^{(t)}}{d_i^{(t)}} < \frac{\tau_{\mathrm{org}}}{2}],
\label{eq:budget_prune}
\end{equation}
where $\tau_{\alpha} = 0.05$ and $\tau_{\mathrm{org}}$ is the base gradient threshold.
Specifically, a primitive is flagged for removal only if it resides in an over-budgeted region, has low opacity, and exhibits negligible gradient magnitude. We integrate this budget-aware constraint with the base pruning mask $P_i^{(t)}$:
\begin{equation}
\tilde P_i^{(t)} =P_i^{(t)} \lor \hat P_i^{(t)}.
\label{eq:prune_fusion}
\end{equation}
To avoid removing low-opacity primitives immediately after an opacity reset, we disable this branch for a cooldown window equal to the standard pruning interval. In addition, newly generated primitives are protected from budget-aware pruning within the same structural update.

\vspace{0.5em}
\noindent\textbf{Equilibrium-driven Termination.}
As the optimization approaches a stable allocation state motivated by the KKT-inspired condition, continuing the fixed-iteration training schedule may yield negligible rendering improvements relative to the incurred computational cost. To detect such convergence, we monitor the exponential moving average of the photometric loss $\bar{L}^{(t)}$ and the primitive count $N^{(t)}$ at fixed intervals $\Delta t$. Their relative temporal variations are computed as
\begin{equation}
\left\{ \begin{array}{l}
\Delta {L^{(t)}} = 1 - \frac{{{{\bar L}^{(t)}}}}{{{{\bar L}^{(t - \Delta t)}}}} \\
\Delta {N^{(t)}} = \frac{{{N^{(t)}}}}{{{N^{(t - \Delta t)}}}} - 1 
\end{array} \right. .
\label{eq:delta_loss_num}
\end{equation}

If both the photometric loss and primitive count remain stable, i.e., $0 \leq \Delta L^{(t)} < \epsilon_L$ and $|\Delta N^{(t)}| < \epsilon_N$, for $\ell$ consecutive checking intervals, the optimization is considered stable and is adaptively terminated. In our implementation, we set $\epsilon_L=0.008$, $\epsilon_N=0.01$, and $\ell=2$. The complete optimization pipeline is summarized in Algorithm~\ref{alg:budget}.

%%%%%%%%%%%%%%%%%%%%%%%%%%%%%%%%%%%%%%%%%%%%%%%%%%%%%%%%%%%%%%%%%%%%%%%%%%%%%%%%%%%%%%%%%%
\begin{algorithm}[t]
\caption{Optimization Process of SPARE-GS}
\label{alg:budget}
\begin{algorithmic}[1]
\REQUIRE A base 3DGS pipeline, training views $\mathcal{V}$, checking interval $\Delta t$, and patience $\ell$
\STATE Initialize base 3D Gaussian primitives
\FOR{optimization step $t=1,\ldots,T$}
    \STATE Render view $v \in \mathcal{V}$, update parameters, and accumulate stats $a_i^{(t)}$, $d_i^{(t)}$, $\rho_i^{(t)}$
    
    \IF{structural evolution (densification/pruning) is triggered}
        \vspace{0.2em}
        \STATE \texttt{\% Part 1: Marginal Utility Proxy (Sec.~\ref{sec4.1})}
        \STATE Discretize spatial volume into regions $\Omega_r$ and map primitives to region indices $k_i^{(t)}$
        \STATE Compute regional signals $H_r^{(t)}, V_r^{(t)}$ and demand proxy $D_r^{(t)}$ \COMMENT{Eqs.~(\ref{eq:demand})--(\ref{eq:region_area})}
        
        \vspace{0.2em}
        \STATE \texttt{\% Part 2: Feedback Control via Target Allocation (Sec.~\ref{sec4.2})}
        \STATE Distribute the global budget according to the capacity-adjusted demand to obtain $\hat{n}_r^{(t)}$
        \COMMENT{Eqs.~(\ref{eq:budget_prob})--(\ref{eq:quota})}
        \STATE Compute structural slack $s_{k_i^{(t)}}^{(t)}$ and budget-aware modulation weight $w_i^{(t)}$ \COMMENT{Eqs.~(\ref{eq:slack}), (\ref{eq:budget_weight})}
        
        \vspace{0.2em}
        \STATE \texttt{\% Part 3: Budget-Guided Structural Modulation (Sec.~\ref{sec4.3})}
        \STATE Modulate the native densification score to obtain $\tilde{S}_i^{(t)}$ 
        \COMMENT{Eq.~(\ref{eq:ours_score})}
        \STATE Execute densification operations based on the modulated score $\tilde{S}_i^{(t)}$
        \STATE Execute redundancy-aware pruning based on budget constraints \COMMENT{Eqs.~(\ref{eq:budget_prune}), (\ref{eq:prune_fusion})}
    \ENDIF

    \IF{$t \bmod \Delta t = 0$}
        \STATE \textbf{\textit{\% Part 3 (Cont.): Equilibrium-Driven Termination (Sec.~\ref{sec4.3})}}
        \STATE Evaluate convergence indicators $\Delta L^{(t)}$ and $\Delta N^{(t)}$ \COMMENT{Eq.~(\ref{eq:delta_loss_num})}
        \IF{the loss does not worsen and both indicators remain stable for $\ell$ consecutive intervals}
            \STATE \textbf{break} \COMMENT{Adaptive termination}
        \ENDIF
    \ENDIF
\ENDFOR
\RETURN Optimized structurally-parsimonious representation
\end{algorithmic}
\end{algorithm}

%%%%%%%%%%%%%%%%%%%%%%%%%%%%%%%%%%%%%%%%%%%%%%%%%%%%%%%%%%%%%%%%%%%%%%%%%%%%%%%%%%%%%%%

\section{Experiments} \label{sec5}
To evaluate the effectiveness and plug-and-play capability of SPARE-GS, we conduct extensive experiments covering training-stage efficiency, downstream compression, and post-training pruning. We further provide in-depth analyses of the key components and hyperparameters, computational overhead, structural evolution, and representative failure cases to better understand the robustness of the proposed framework.

\subsection{Experiment Settings}
\noindent \textbf{Datasets.} Following the evaluation of the original 3D-GS~\cite{kerbl20233d}, we tested our SPARE-GS on the same challenging real-world scenes. Specifically, we used all nine scenes from the Mip-NeRF 360 dataset~\cite{barron2022mipnerf360}, which contains five outdoor and four indoor scenes, each featuring complex central objects or viewing areas and detailed backgrounds. Additionally, two outdoor scenes, $truck$ and $train$, were taken from the Tanks \& Temples dataset~\cite{knapitsch2017tanks}, and two indoor scenes, $drjohnson$ and $playroom$, were taken from the Deep Blending dataset~\cite{hedman2018deep}. Furthermore, we supplemented our evaluation with eight urban scenes from the BungeeNeRF dataset~\cite{xiangli2022bungeenerf} to comprehensively assess our method. For consistency, we used the COLMAP camera pose estimates provided in the original 3D-GS~\cite{kerbl20233d}. 

\vspace{0.5em}
\noindent\textbf{Implementation Details.}
Our SPARE-GS is a plug-and-play, purely online structural regulation framework that can be integrated into the existing 3DGS training pipeline. To highlight the effectiveness of the budget-guided mechanism itself, all integrated comparisons regarding structural efficiency were conducted under a \textit{drop-in} condition. That is, rather than altering the specific densification and pruning strategies of the underlying methods, we solely applied our modulation to their native heuristics, with no pipeline-specific hyperparameter tuning.

%%%%%%%%%%%%%%%%%%%%%%%%%%%%%%%%%%%%%%%%%%%%%%%%%%%%%%%%%%%%%%%%%%%%%%
\vspace{0.5em}
\noindent \textbf{Evaluation Metrics.} 
To ensure a fair comparison, all baselines and our adapted versions were executed on the same hardware setup (Intel i9 CPU, RTX 3090 GPU). We evaluated performance using three standard image quality metrics: Peak Signal-to-Noise Ratio (PSNR), Structural Similarity (SSIM)~\cite{wang2004image}, and Learned Perceptual Image Patch Similarity (LPIPS)~\cite{zhang2018unreasonable}. Beyond these, we validated our resource-constrained formulation by tracking the total training time and the active Gaussian primitive count. For downstream tasks, we further reported the Bjøntegaard Delta (BD) metrics to measure rate-distortion efficiency, alongside coding times to quantify the associated computational overhead.

\begin{table*}[htbp] % 增加排版引擎的浮动灵活性
\centering
\caption{\textbf{Quantitative evaluation of our method in terms of 3DGS training efficiency across different base pipelines and datasets.} $\boldsymbol{\varDelta}$ indicates the relative or absolute difference between the base method and our SPARE-GS. Bold values indicate improvements or resource reductions. $\boldsymbol{\bar{\varDelta}}$ denotes the scene-count-weighted mean change.}
\label{tab:comprehensive_results}
\renewcommand{\arraystretch}{0.80} 
% ==========================================

\resizebox{\textwidth}{!}{
\begin{tabular}{@{}lllccccc@{}}
\toprule
\textbf{Base Pipeline} & \textbf{Dataset} & \textbf{Method} & \textbf{Time (s)} $\downarrow$ & \textbf{Gaussians} $\downarrow$ & \textbf{PSNR} $\uparrow$ & \textbf{SSIM} $\uparrow$ & \textbf{LPIPS} $\downarrow$ \\

\midrule
% ==================== 1. Vanilla 3DGS ====================
\multirow{12}{*}{\textbf{Vanilla 3DGS}~\cite{kerbl20233d}} 
 & \multirow{3}{*}{Mip-NeRF 360~\cite{barron2022mipnerf360}} & Baseline & 1327.08 & 2,640,494 & 27.52 & 0.813 & 0.221 \\
 & & + SPARE-GS (Ours) & 939.23 & 1,691,678 & 27.44 & 0.809 & 0.234 \\
 & & \cellcolor{gray!20}$\boldsymbol{\varDelta}$ & \cellcolor{gray!20}\textit{\textbf{-29.23\%}} & \cellcolor{gray!20}\textit{\textbf{-35.93\%}} & \cellcolor{gray!20}\textit{-0.08} & \cellcolor{gray!20}\textit{-0.004} & \cellcolor{gray!20}\textit{+0.013} \\
\cmidrule{2-8}
 & \multirow{3}{*}{Deep Blending~\cite{hedman2018deep}} & Baseline & 1257.89 & 2,458,697 & 29.63 & 0.902 & 0.241 \\
 & & + SPARE-GS (Ours) & 779.11 & 1,124,510 & 29.77 & 0.906 & 0.246 \\
 & & \cellcolor{gray!20}$\boldsymbol{\varDelta}$ & \cellcolor{gray!20}\textit{\textbf{-38.06\%}} & \cellcolor{gray!20}\textit{\textbf{-54.26\%}} & \cellcolor{gray!20}\textit{\textbf{+0.14}} & \cellcolor{gray!20}\textit{\textbf{+0.004}} & \cellcolor{gray!20}\textit{+0.005} \\
\cmidrule{2-8}
 & \multirow{3}{*}{Tanks \& Temples~\cite{knapitsch2017tanks}} & Baseline & 718.15 & 1,572,798 & 23.83 & 0.850 & 0.171 \\
 & & + SPARE-GS (Ours) & 485.66 & 973,619 & 24.01 & 0.849 & 0.181 \\
 & & \cellcolor{gray!20}$\boldsymbol{\varDelta}$ & \cellcolor{gray!20}\textit{\textbf{-32.37\%}} & \cellcolor{gray!20}\textit{\textbf{-38.10\%}} & \cellcolor{gray!20}\textit{\textbf{+0.18}} & \cellcolor{gray!20}\textit{-0.001} & \cellcolor{gray!20}\textit{+0.010} \\
\cmidrule{2-8}
 & \multirow{3}{*}{BungeeNeRF~\cite{xiangli2022bungeenerf}} & Baseline & 2364.79 & 6,506,221 & 27.83 & 0.915 & 0.098 \\
 & & + SPARE-GS (Ours) & 1478.60 & 3,589,073 & 27.74 & 0.916 & 0.106 \\
 & & \cellcolor{gray!20}$\boldsymbol{\varDelta}$ & \cellcolor{gray!20}\textit{\textbf{-37.47\%}} & \cellcolor{gray!20}\textit{\textbf{-44.84\%}} & \cellcolor{gray!20}\textit{-0.09} & \cellcolor{gray!20}\textit{\textbf{+0.001}} & \cellcolor{gray!20}\textit{+0.008} \\

\midrule
% ==================== 2. Mip-Splatting ====================
\multirow{12}{*}{\textbf{Mip-Splatting}~\cite{yu2024mip}} 
 & \multirow{3}{*}{Mip-NeRF 360~\cite{barron2022mipnerf360}} & Baseline & 2727.37 & 4,202,055 & 27.74 & 0.827 & 0.190 \\
 & & + SPARE-GS (Ours) & 1899.76 & 2,511,604 & 27.72 & 0.826 & 0.201 \\
 & & \cellcolor{gray!20}$\boldsymbol{\varDelta}$ & \cellcolor{gray!20}\textit{\textbf{-30.34\%}} & \cellcolor{gray!20}\textit{\textbf{-40.23\%}} & \cellcolor{gray!20}\textit{-0.02} & \cellcolor{gray!20}\textit{-0.001} & \cellcolor{gray!20}\textit{+0.011} \\
\cmidrule{2-8}
 & \multirow{3}{*}{Deep Blending~\cite{hedman2018deep}} & Baseline & 2246.00 & 3,503,352 & 29.43 & 0.903 & 0.239 \\
 & & + SPARE-GS (Ours) & 1367.55 & 1,752,417 & 29.75 & 0.908 & 0.240 \\
 & & \cellcolor{gray!20}$\boldsymbol{\varDelta}$ & \cellcolor{gray!20}\textit{\textbf{-39.11\%}} & \cellcolor{gray!20}\textit{\textbf{-49.98\%}} & \cellcolor{gray!20}\textit{\textbf{+0.32}} & \cellcolor{gray!20}\textit{\textbf{+0.005}} & \cellcolor{gray!20}\textit{+0.001} \\
\cmidrule{2-8}
 & \multirow{3}{*}{Tanks \& Temples~\cite{knapitsch2017tanks}} & Baseline & 1415.70 & 2,358,386 & 23.80 & 0.860 & 0.157 \\
 & & + SPARE-GS (Ours) & 946.15 & 1,371,594 & 24.10 & 0.860 & 0.165 \\
 & & \cellcolor{gray!20}$\boldsymbol{\varDelta}$ & \cellcolor{gray!20}\textit{\textbf{-33.17\%}} & \cellcolor{gray!20}\textit{\textbf{-41.84\%}} & \cellcolor{gray!20}\textit{\textbf{+0.30}} & \cellcolor{gray!20}\textit{0.000} & \cellcolor{gray!20}\textit{+0.008} \\
\cmidrule{2-8}
 & \multirow{3}{*}{BungeeNeRF~\cite{xiangli2022bungeenerf}} & Baseline & 4015.86 & 7,675,877 & 28.25 & 0.924 & 0.088 \\
 & & + SPARE-GS (Ours) & 2604.42 & 4,336,878 & 28.06 & 0.921 & 0.097 \\
 & & \cellcolor{gray!20}$\boldsymbol{\varDelta}$ & \cellcolor{gray!20}\textit{\textbf{-35.15\%}} & \cellcolor{gray!20}\textit{\textbf{-43.50\%}} & \cellcolor{gray!20}\textit{-0.19} & \cellcolor{gray!20}\textit{-0.003} & \cellcolor{gray!20}\textit{+0.009} \\
 
\midrule
% ==================== 3. TamingGS ====================
\multirow{12}{*}{\textbf{TamingGS}~\cite{mallick2024taming}} 
 & \multirow{3}{*}{Mip-NeRF 360~\cite{barron2022mipnerf360}} & Baseline & 871.08 & 1,416,946 & 27.82 & 0.805 & 0.240 \\
 & & + SPARE-GS (Ours) & 716.64 & 1,052,388 & 27.77 & 0.808 & 0.238 \\
 & & \cellcolor{gray!20}$\boldsymbol{\varDelta}$ & \cellcolor{gray!20}\textit{\textbf{-17.73\%}} & \cellcolor{gray!20}\textit{\textbf{-25.73\%}} & \cellcolor{gray!20}\textit{-0.05} & \cellcolor{gray!20}\textit{\textbf{+0.003}} & \cellcolor{gray!20}\textit{\textbf{-0.002}} \\
\cmidrule{2-8}
 & \multirow{3}{*}{Deep Blending~\cite{hedman2018deep}} & Baseline & 597.56 & 1,178,605 & 29.97 & 0.907 & 0.246 \\
 & & + SPARE-GS (Ours) & 550.27 & 1,052,175 & 30.11 & 0.911 & 0.243 \\
 & & \cellcolor{gray!20}$\boldsymbol{\varDelta}$ & \cellcolor{gray!20}\textit{\textbf{-7.91\%}} & \cellcolor{gray!20}\textit{\textbf{-10.73\%}} & \cellcolor{gray!20}\textit{\textbf{+0.14}} & \cellcolor{gray!20}\textit{\textbf{+0.004}} & \cellcolor{gray!20}\textit{\textbf{-0.003}} \\
\cmidrule{2-8}
 & \multirow{3}{*}{Tanks \& Temples~\cite{knapitsch2017tanks}} & Baseline & 848.80 & 2,206,907 & 24.47 & 0.863 & 0.156 \\
 & & + SPARE-GS (Ours) & 518.31 & 981,097 & 24.41 & 0.860 & 0.167 \\
 & & \cellcolor{gray!20}$\boldsymbol{\varDelta}$ & \cellcolor{gray!20}\textit{\textbf{-38.94\%}} & \cellcolor{gray!20}\textit{\textbf{-55.54\%}} & \cellcolor{gray!20}\textit{-0.06} & \cellcolor{gray!20}\textit{-0.003} & \cellcolor{gray!20}\textit{+0.011} \\
\cmidrule{2-8}
 & \multirow{3}{*}{BungeeNeRF~\cite{xiangli2022bungeenerf}} & Baseline & 1209.50 & 2,426,465 & 27.39 & 0.903 & 0.127 \\
 & & + SPARE-GS (Ours) & 1210.51 & 2,422,162 & 27.71 & 0.913 & 0.113 \\
 & & \cellcolor{gray!20}$\boldsymbol{\varDelta}$ & \cellcolor{gray!20}\textit{+0.08\%} & \cellcolor{gray!20}\textit{\textbf{-0.18\%}} & \cellcolor{gray!20}\textit{\textbf{+0.32}} & \cellcolor{gray!20}\textit{\textbf{+0.010}} & \cellcolor{gray!20}\textit{\textbf{-0.014}} \\

\midrule
% ==================== 4. DashGaussian ====================
\multirow{12}{*}{\textbf{DashGaussian}~\cite{chen2025dashgaussian}} 
 & \multirow{3}{*}{Mip-NeRF 360~\cite{barron2022mipnerf360}} & Baseline & 1102.76 & 2,532,764 & 27.73 & 0.815 & 0.220 \\
 & & + SPARE-GS (Ours) & 803.87 & 1,762,803 & 27.65 & 0.814 & 0.227 \\
 & & \cellcolor{gray!20}$\boldsymbol{\varDelta}$ & \cellcolor{gray!20}\textit{\textbf{-27.10\%}} & \cellcolor{gray!20}\textit{\textbf{-30.40\%}} & \cellcolor{gray!20}\textit{-0.08} & \cellcolor{gray!20}\textit{-0.001} & \cellcolor{gray!20}\textit{+0.007} \\
\cmidrule{2-8}
 & \multirow{3}{*}{Deep Blending~\cite{hedman2018deep}} & Baseline & 858.09 & 2,313,205 & 29.60 & 0.906 & 0.237 \\
 & & + SPARE-GS (Ours) & 579.23 & 1,129,622 & 29.78 & 0.907 & 0.241 \\
 & & \cellcolor{gray!20}$\boldsymbol{\varDelta}$ & \cellcolor{gray!20}\textit{\textbf{-32.50\%}} & \cellcolor{gray!20}\textit{\textbf{-51.17\%}} & \cellcolor{gray!20}\textit{\textbf{+0.18}} & \cellcolor{gray!20}\textit{\textbf{+0.001}} & \cellcolor{gray!20}\textit{+0.004} \\
\cmidrule{2-8}
 & \multirow{3}{*}{Tanks \& Temples~\cite{knapitsch2017tanks}} & Baseline & 698.44 & 1,501,090 & 24.15 & 0.858 & 0.165 \\
 & & + SPARE-GS (Ours) & 498.90 & 931,448 & 24.39 & 0.857 & 0.175 \\
 & & \cellcolor{gray!20}$\boldsymbol{\varDelta}$ & \cellcolor{gray!20}\textit{\textbf{-28.57\%}} & \cellcolor{gray!20}\textit{\textbf{-37.95\%}} & \cellcolor{gray!20}\textit{\textbf{+0.24}} & \cellcolor{gray!20}\textit{-0.001} & \cellcolor{gray!20}\textit{+0.010} \\
\cmidrule{2-8}
 & \multirow{3}{*}{BungeeNeRF~\cite{xiangli2022bungeenerf}} & Baseline & 2266.32 & 5,054,017 & 27.94 & 0.918 & 0.095 \\
 & & + SPARE-GS (Ours) & 1451.74 & 2,994,782 & 28.10 & 0.924 & 0.095 \\
 & & \cellcolor{gray!20}$\boldsymbol{\varDelta}$ & \cellcolor{gray!20}\textit{\textbf{-35.94\%}} & \cellcolor{gray!20}\textit{\textbf{-40.74\%}} & \cellcolor{gray!20}\textit{\textbf{+0.16}} & \cellcolor{gray!20}\textit{\textbf{+0.006}} & \cellcolor{gray!20}\textit{0.000} \\

\midrule
% ==================== 5. FastGS ====================
\multirow{12}{*}{\textbf{FastGS}~\cite{ren2025fastgs}} 
 & \multirow{3}{*}{Mip-NeRF 360~\cite{barron2022mipnerf360}} & Baseline & 269.08 & 558,202 & 27.29 & 0.800 & 0.254 \\
 & & + SPARE-GS (Ours) & 248.83 & 508,898 & 27.26 & 0.797 & 0.259 \\
 & & \cellcolor{gray!20}$\boldsymbol{\varDelta}$ & \cellcolor{gray!20}\textit{\textbf{-7.53\%}} & \cellcolor{gray!20}\textit{\textbf{-8.83\%}} & \cellcolor{gray!20}\textit{-0.03} & \cellcolor{gray!20}\textit{-0.003} & \cellcolor{gray!20}\textit{+0.005} \\
\cmidrule{2-8}
 & \multirow{3}{*}{Deep Blending~\cite{hedman2018deep}} & Baseline & 168.76 & 316,890 & 30.14 & 0.908 & 0.254 \\
 & & + SPARE-GS (Ours) & 158.02 & 281,643 & 30.09 & 0.908 & 0.256 \\
 & & \cellcolor{gray!20}$\boldsymbol{\varDelta}$ & \cellcolor{gray!20}\textit{\textbf{-6.36\%}} & \cellcolor{gray!20}\textit{\textbf{-11.12\%}} & \cellcolor{gray!20}\textit{-0.05} & \cellcolor{gray!20}\textit{0.000} & \cellcolor{gray!20}\textit{+0.002} \\
\cmidrule{2-8}
 & \multirow{3}{*}{Tanks \& Temples~\cite{knapitsch2017tanks}} & Baseline & 151.04 & 247,905 & 23.48 & 0.837 & 0.212 \\
 & & + SPARE-GS (Ours) & 139.74 & 221,914 & 23.46 & 0.834 & 0.216 \\
 & & \cellcolor{gray!20}$\boldsymbol{\varDelta}$ & \cellcolor{gray!20}\textit{\textbf{-7.48\%}} & \cellcolor{gray!20}\textit{\textbf{-10.48\%}} & \cellcolor{gray!20}\textit{-0.02} & \cellcolor{gray!20}\textit{-0.003} & \cellcolor{gray!20}\textit{+0.004} \\
\cmidrule{2-8}
 & \multirow{3}{*}{BungeeNeRF~\cite{xiangli2022bungeenerf}} & Baseline & 453.17 & 1,173,619 & 27.30 & 0.903 & 0.140 \\
 & & + SPARE-GS (Ours) & 399.10 & 933,641 & 27.21 & 0.895 & 0.148 \\
 & & \cellcolor{gray!20}$\boldsymbol{\varDelta}$ & \cellcolor{gray!20}\textit{\textbf{-11.93\%}} & \cellcolor{gray!20}\textit{\textbf{-20.45\%}} & \cellcolor{gray!20}\textit{-0.09} & \cellcolor{gray!20}\textit{-0.008} & \cellcolor{gray!20}\textit{+0.008} \\

\midrule\midrule
% ==================== Overall Mean Change ====================
\multicolumn{2}{@{}l}{\textbf{Average}} & \cellcolor{gray!20}$\boldsymbol{\bar{\varDelta}}$ & \cellcolor{gray!20}\textit{\textbf{-23.81\%}} & \cellcolor{gray!20}\textit{\textbf{-30.38\%}} & \cellcolor{gray!20}\textit{\textbf{+0.01}} & \cellcolor{gray!20}\textit{0.000} & \cellcolor{gray!20}\textit{+0.005} \\
 
\bottomrule
\end{tabular}
}
\end{table*}

\begin{figure*}[]
  \centering
  \includegraphics[width=0.96\linewidth]{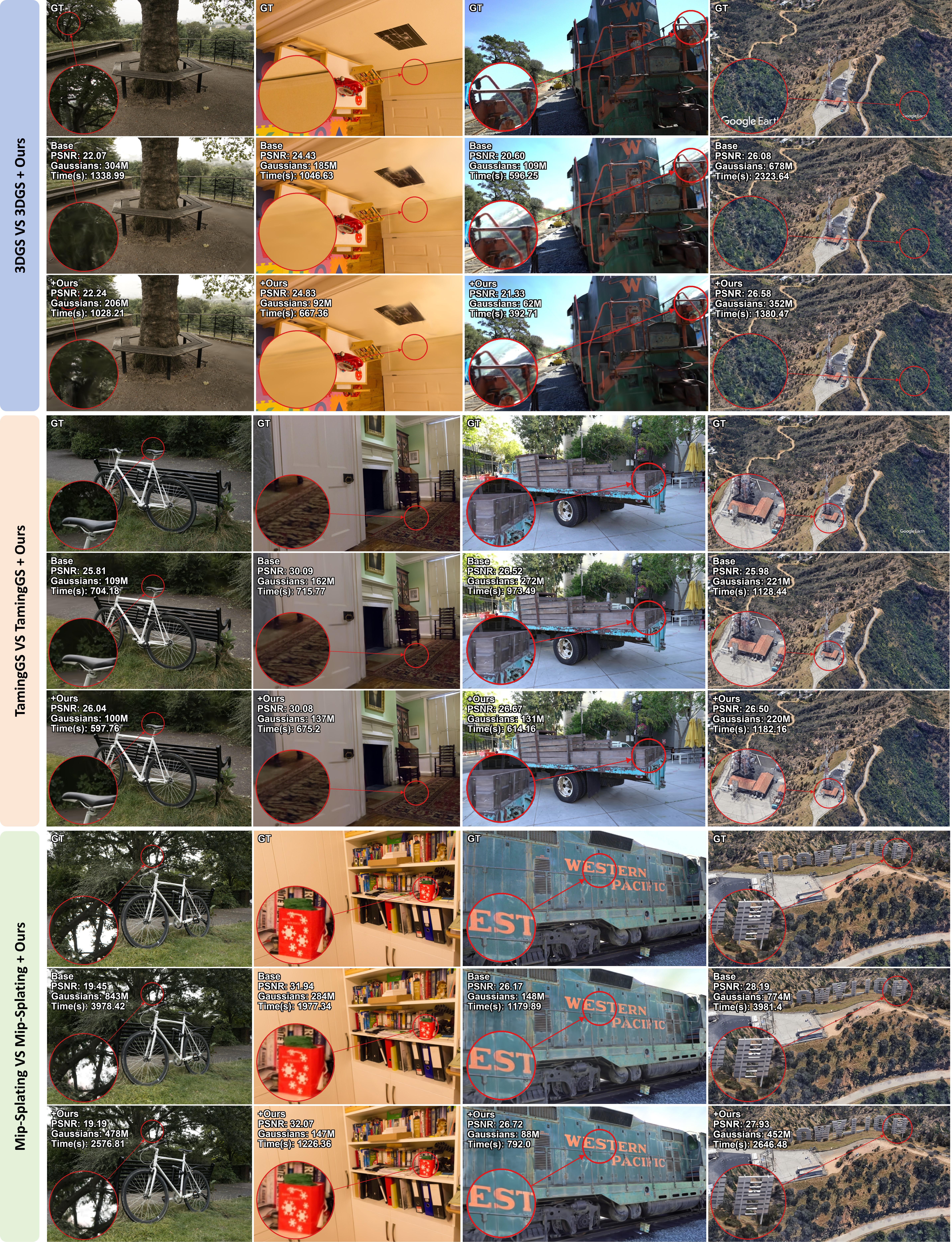}
\caption{\textbf{Comparison of rendering quality.} {Rows (Top to Bottom):} Grouped by base methods into 3DGS, TamingGS, and Mip-Splatting. Within each three-row block, we sequentially show the Ground Truth (GT), the baseline method (Base), and the baseline enhanced by our method (\textbf{+ Ours}). {Columns (Left to Right):} Representative scenes from Mip-NeRF 360, Deep Blending, Tanks \& Temples, and BungeeNeRF. Please zoom in for details.}
\label{fig:visual_result1}
\end{figure*}

\begin{figure*}[]
  \centering
  \includegraphics[width=0.96\linewidth]{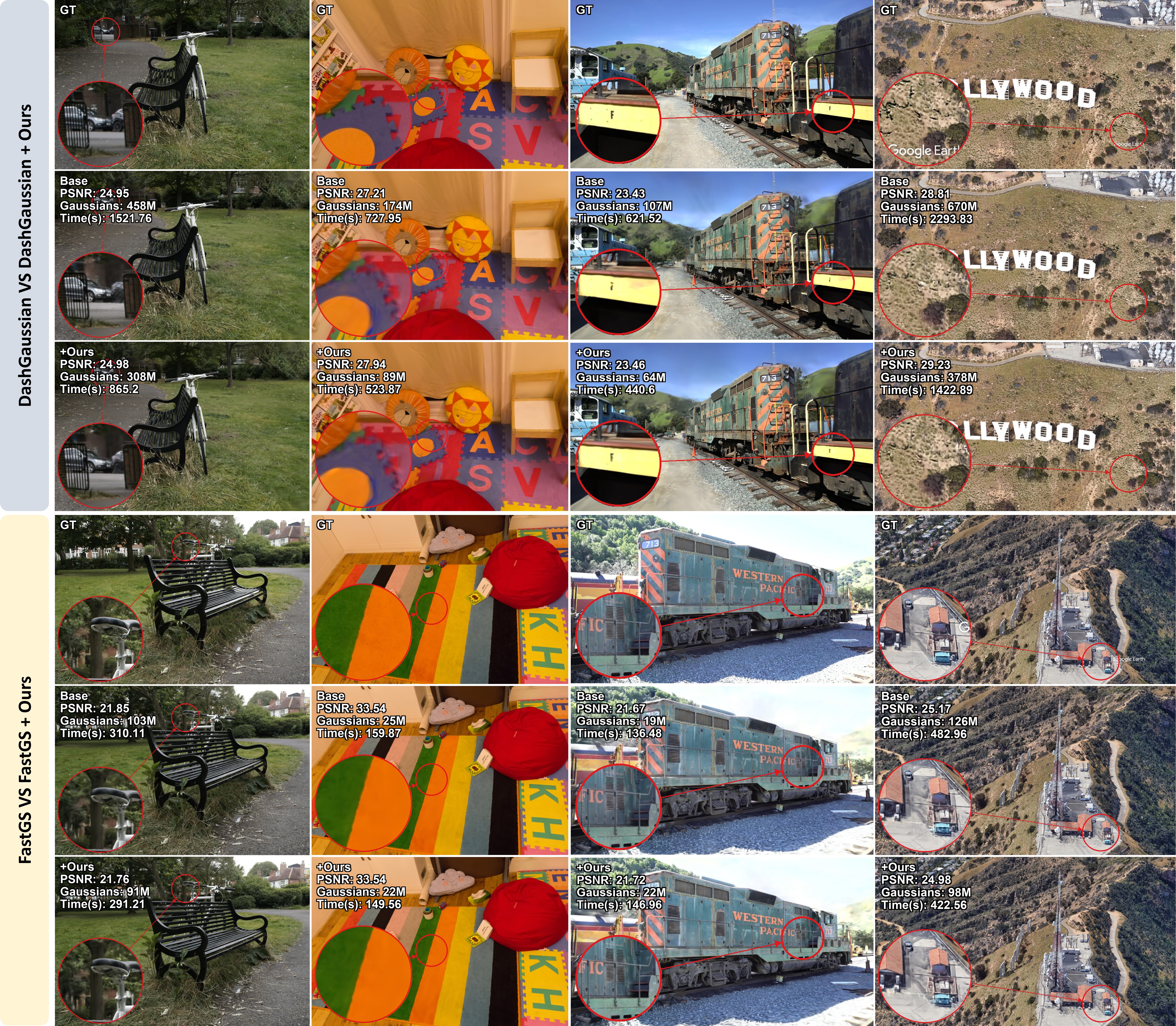}
\caption{\textbf{Comparison of rendering quality.} {Rows (Top to Bottom):} Grouped by base methods into DashGaussian and FastGS. Within each three-row block, we sequentially show the Ground Truth (GT), the baseline method (Base), and the baseline enhanced by our method (\textbf{+ Ours}). {Columns (Left to Right):} Representative scenes from Mip-NeRF 360, Deep Blending, Tanks \& Temples, and BungeeNeRF. Please zoom in for details.}
\label{fig:visual_result2}
\end{figure*}

\subsection{Evaluation on 3DGS Training Efficiency}

In this section, we evaluated the efficacy of SPARE-GS in optimizing the efficiency-fidelity trade-off and constraining the final primitive footprint. To substantiate its generalizability, we integrated our framework into five representative pipelines by augmenting their default density control mechanisms with our budget-guided structural modulation. The evaluated pipelines include the original/vanilla  3DGS~\cite{kerbl20233d}, a quality-enhanced variant (Mip-Splatting~\cite{yu2024mip}), and three acceleration-oriented frameworks (TamingGS~\cite{mallick2024taming}, DashGaussian~\cite{chen2025dashgaussian}, and FastGS~\cite{ren2025fastgs}).
Table~\ref{tab:comprehensive_results} reports the training time, final Gaussian count, and rendering fidelity in terms of PSNR, SSIM, and LPIPS. Furthermore, qualitative comparisons demonstrating these structural benefits are provided in Figs.~\ref{fig:visual_result1} and~\ref{fig:visual_result2}.

\vspace{0.5em}
\noindent\textbf{Performance on Vanilla 3DGS~\cite{kerbl20233d}.}
The results on the original 3DGS support our analysis regarding structural redundancy. Relying on local gradient heuristics, the standard pipeline tends to over-allocate primitives in high-frequency regions. By applying our budget-guided structural modulation and adaptive termination, we reduce the final Gaussian primitive count by 35.93\% to 54.26\% and decrease the total training time by 29.23\% to 38.06\%. This reduction in structural complexity is achieved while maintaining comparable rendering quality. On the Deep Blending and Tanks \& Temples datasets, the PSNR shows slight improvements (+0.14 and +0.18 dB, respectively), while on Mip-NeRF 360 and BungeeNeRF, the decreases are marginal (-0.08 and -0.09 dB, respectively). These results suggest that many suppressed primitives provide
limited additional contribution to reconstruction quality, which is consistent with our budget-constrained motivation.

\vspace{0.5em}
\noindent\textbf{Performance on Mip-Splatting~\cite{yu2024mip}.} 
Mip-Splatting improves alias-free rendering by augmenting 3DGS with scale-aware filtering mechanisms.
Table~\ref{tab:comprehensive_results} demonstrates a similar trend when SPARE-GS is integrated into this pipeline. Across all four datasets, including the large-scale BungeeNeRF, our method substantially reduces the Gaussian count by 40.23\% to 49.98\% and accelerates training by 30.34\% to 39.11\%. Notably, the quality response varies across datasets. PSNR improves by
0.32 dB and 0.30 dB on Deep Blending and Tanks \& Temples, respectively, while decreasing by 0.02 dB on Mip-NeRF 360 and 0.19 dB on BungeeNeRF. 

\vspace{0.5em}
\noindent\textbf{Performance on TamingGS~\cite{mallick2024taming}.} 
TamingGS aims to construct high-quality radiance fields under limited resources by introducing targeted structural allocation. As shown in Table~\ref{tab:comprehensive_results}, despite this resource-aware baseline, integrating SPARE-GS further reduces the number of Gaussian primitives by 10.73\% to 55.54\% and decreases the training time by 7.91\% to 38.94\% across Mip-NeRF 360, Deep Blending, and Tanks \& Temples. These resource reductions are achieved with only limited variations in rendering quality. The PSNR variation remains minimal, bounded within $\pm$0.06 dB on Mip-NeRF 360 and Tanks \& Temples, while Deep Blending achieves a 0.14 dB improvement. Furthermore, on BungeeNeRF, the Gaussian count and training time remain virtually unchanged,
with changes of -0.18\% and +0.08\%, respectively, while PSNR, SSIM, and LPIPS improve by 0.32 dB, 0.010, and 0.014, respectively. This result indicates that SPARE-GS does not enforce uniform compression; instead, it adapts the
structural allocation to the demand of the underlying scene and pipeline.

\vspace{0.5em}
\noindent\textbf{Performance on DashGaussian~\cite{chen2025dashgaussian}.}
DashGaussian accelerates optimization through dynamic rendering resolutions and coordinated primitive growth. As shown in Table~\ref{tab:comprehensive_results}, integrating our SPARE-GS further reduces the training time by 27.10\% to 35.94\% and decreases the final primitive count by 30.40\% to 51.17\%, while maintaining competitive reconstruction quality. Although DashGaussian controls the number of selected candidates at each densification step through global top-(k) ranking, it does not explicitly account for region-wise budget imbalance. SPARE-GS complements this mechanism by incorporating regional budget signals into the native ranking score, thereby increasing the priority of candidates located in under-budgeted regions without altering the original candidate-generation procedure.

\vspace{0.5em}
\noindent\textbf{Performance on FastGS~\cite{ren2025fastgs}.}
FastGS accelerates training by incorporating a multi-view consistency score, which functions as a structural regularizer and yields highly compact scene representations (e.g., producing $\sim$558K primitives on Mip-NeRF 360). Table~\ref{tab:comprehensive_results} demonstrates that our budget-guided framework successfully identifies and eliminates residual redundancies in this baseline. SPARE-GS further reduces the Gaussian count by 8.83\% to 20.45\% and the training time by 6.36\% to 11.93\%. Since FastGS already produces highly compact representations, the additional efficiency gains are comparatively smaller than those observed on other pipelines. The PSNR variations remain limited, ranging from -0.09 dB to -0.02 dB.

\vspace{0.5em}
\noindent\textbf{Overall.} 
Across different pipeline configurations, SPARE-GS reduces the Gaussian count and training time by an average of 30.38\% and 23.81\%, respectively. These efficiency gains are accompanied by an average PSNR improvement of 0.01 dB and a negligible mean SSIM change, while LPIPS increases slightly by 0.005. The magnitude of the gain varies with the structural characteristics of each base pipeline: substantial reductions are observed on Vanilla 3DGS, Mip-Splatting, and DashGaussian, whereas the improvements are more moderate on the already compact FastGS pipeline. On TamingGS with BungeeNeRF, SPARE-GS preserves essentially the same resource consumption while noticeably improving rendering metrics. Overall, these results demonstrate that the proposed budget-guided regulation adapts to different structural regimes and generally provides a more favorable efficiency-quality trade-off.

%%%%%%%%%%%%%%%%%%%%%%%%%%%%%%%%%%%%%%%%%%%%%%%%%%%%%%%%%%%%%%%%%%%%%%%

\subsection{Evaluation on 3DGS Post-Processing Performance}
\begin{table*}[htbp]
\centering
\caption{\textbf{Quantitative evaluation of our method in terms of 3DGS post-processing compression.} The rightmost columns report relative processing time reductions ($\Delta$Time) and Bjøntegaard Delta (BD) metrics. `N/A' indicates insufficient overlap between the corresponding R-D curves to compute the reported BD metric. }

\label{tab:compression_perspective}
\setlength{\tabcolsep}{2.5pt} 
\renewcommand{\arraystretch}{0.85} 
\setlength{\aboverulesep}{2pt} 
\setlength{\belowrulesep}{2pt} 
\scriptsize 
\resizebox{\textwidth}{!}{
\begin{tabular}{@{}l ccccc ccccc ccc@{}}
\toprule
\multirow{2}{*}{\textbf{Dataset}} & \multicolumn{5}{c}{\textbf{Input: Vanilla 3DGS (Base)}} & \multicolumn{5}{c}{\textbf{\textbf{Input: SPARE-GS (Ours)}}} & \multicolumn{3}{c}{\textbf{Ours vs. Base}} \\
\cmidrule(lr){2-6} \cmidrule(lr){7-11} \cmidrule(l){12-14}
& \textbf{Size (MB)} & \textbf{Time (s)} & \textbf{PSNR} & \textbf{SSIM} & \textbf{LPIPS} & \textbf{Size (MB)} & \textbf{Time (s)} & \textbf{PSNR} & \textbf{SSIM} & \textbf{LPIPS} & \textbf{$\Delta$Time} $\downarrow$ & \textbf{BD-Rate} $\downarrow$ & \textbf{BD-PSNR} $\uparrow$ \\

\midrule
% ================= Block: LightGaussian =================
\multicolumn{14}{@{}l}{\cellcolor{gray!20}\textbf{Compression Method: LightGaussian}~\cite{fan2024lightgaussian}} \\
\multirow{3}{*}{\textbf{Mip-NeRF 360}} 
& 407.56 & 101.25 & 27.19 & 0.807 & 0.228 & 244.08 & 85.89 & 27.05 & 0.802 & 0.244 & \multirow{3}{*}{\textbf{-13.89\%}} & \multirow{3}{*}{\textbf{-27.78\%}} & \multirow{3}{*}{N/A} \\
& 321.67 & 97.40 & 26.98 & 0.802 & 0.233 & 192.69 & 83.72 & 26.83 & 0.798 & 0.248 & & & \\
& 235.90 & 92.23 & 26.76 & 0.797 & 0.239 & 141.19 & 80.87 & 26.58 & 0.792 & 0.254 & & & \\
\cmidrule{2-14}
\multirow{3}{*}{\textbf{Deep Blending}} 
& 365.53 & 98.21 & 29.66 & 0.905 & 0.242 & 150.59 & 77.36 & 29.63 & 0.908 & 0.250 & \multirow{3}{*}{\textbf{-19.16\%}} & \multirow{3}{*}{\textbf{-55.38\%}} & \multirow{3}{*}{N/A} \\
& 289.13 & 94.27 & 29.54 & 0.903 & 0.244 & 119.31 & 76.77 & 29.48 & 0.906 & 0.252 & & & \\
& 212.78 & 90.75 & 29.31 & 0.900 & 0.247 & 88.09 & 74.82 & 29.30 & 0.904 & 0.255 & & & \\
\cmidrule{2-14}
\multirow{3}{*}{\textbf{Tanks \& Temples}} 
& 238.77 & 86.05 & 23.56 & 0.849 & 0.174 & 137.43 & 76.25 & 23.69 & 0.841 & 0.194 & \multirow{3}{*}{\textbf{-10.42\%}} & \multirow{3}{*}{\textbf{-51.75\%}} & \multirow{3}{*}{N/A} \\
& 188.82 & 83.51 & 23.44 & 0.845 & 0.178 & 108.81 & 75.20 & 23.53 & 0.837 & 0.198 & & & \\
& 138.66 & 81.57 & 23.29 & 0.842 & 0.182 & 79.85 & 73.51 & 23.40 & 0.833 & 0.202 & & & \\
\cmidrule{2-14}
\multirow{3}{*}{\textbf{BungeeNeRF}} 
& 1020.73 & 158.37 & 27.68 & 0.911 & 0.103 & 510.44 & 112.34 & 27.56 & 0.910 & 0.115 & \multirow{3}{*}{\textbf{-28.23\%}} & \multirow{3}{*}{\textbf{-37.50\%}} & \multirow{3}{*}{N/A} \\
& 804.53 & 149.28 & 27.51 & 0.908 & 0.108 & 402.31 & 106.59 & 27.37 & 0.906 & 0.120 & & & \\
& 587.86 & 136.87 & 27.36 & 0.905 & 0.112 & 294.05 & 100.08 & 27.18 & 0.902 & 0.125 & & & \\

\midrule
% ================= Block: SOG =================
\multicolumn{14}{@{}l}{\cellcolor{gray!20}\textbf{Compression Method: SOG}~\cite{morgenstern2024compact}} \\
\multirow{3}{*}{\textbf{Mip-NeRF 360}} 
& 45.92 & 39.45 & 26.44 & 0.782 & 0.255 & 28.13 & 26.63 & 26.25 & 0.777 & 0.271 & \multirow{3}{*}{\textbf{-32.65\%}} & \multirow{3}{*}{\textbf{-35.60\%}} & \multirow{3}{*}{N/A} \\
& 39.28 & 38.64 & 25.98 & 0.772 & 0.267 & 24.13 & 26.01 & 25.80 & 0.766 & 0.283 & & & \\
& 29.56 & 37.46 & 24.71 & 0.742 & 0.298 & 18.18 & 25.18 & 24.49 & 0.736 & 0.314 & & & \\
\cmidrule{2-14}
\multirow{3}{*}{\textbf{Deep Blending}} 
& 45.20 & 36.81 & 29.35 & 0.900 & 0.246 & 19.27 & 19.88 & 29.28 & 0.902 & 0.256 & \multirow{3}{*}{\textbf{-46.77\%}} & \multirow{3}{*}{\textbf{-55.99\%}} & \multirow{3}{*}{N/A} \\
& 38.64 & 35.77 & 29.05 & 0.897 & 0.251 & 16.54 & 18.87 & 28.99 & 0.899 & 0.261 & & & \\
& 28.54 & 34.42 & 28.02 & 0.883 & 0.270 & 12.27 & 18.21 & 27.81 & 0.884 & 0.280 & & & \\
\cmidrule{2-14}
\multirow{3}{*}{\textbf{Tanks \& Temples}} 
& 25.84 & 26.97 & 23.10 & 0.830 & 0.194 & 15.21 & 18.23 & 23.17 & 0.821 & 0.216 & \multirow{3}{*}{\textbf{-32.89\%}} & \multirow{3}{*}{\textbf{-39.88\%}} & \multirow{3}{*}{N/A} \\
& 22.02 & 26.64 & 22.80 & 0.821 & 0.205 & 13.03 & 17.74 & 22.79 & 0.811 & 0.228 & & & \\
& 16.57 & 25.66 & 21.85 & 0.794 & 0.237 & 9.87 & 17.23 & 21.74 & 0.783 & 0.260 & & & \\
\cmidrule{2-14}
\multirow{3}{*}{\textbf{BungeeNeRF}} 
& 128.33 & 80.67 & 27.40 & 0.903 & 0.116 & 66.24 & 45.68 & 27.15 & 0.898 & 0.132 & \multirow{3}{*}{\textbf{-43.74\%}} & \multirow{3}{*}{\textbf{-44.17\%}} & \multirow{3}{*}{N/A} \\
& 110.47 & 77.24 & 27.00 & 0.895 & 0.127 & 57.22 & 43.65 & 26.72 & 0.889 & 0.144 & & & \\
& 82.65 & 74.05 & 25.71 & 0.868 & 0.163 & 42.91 & 41.18 & 25.43 & 0.859 & 0.181 & & & \\

\midrule

% ================= Block: MesonGS =================
\multicolumn{14}{@{}l}{\cellcolor{gray!20}\textbf{Compression Method: MesonGS}~\cite{xie2024mesongs}} \\
\multirow{3}{*}{\textbf{Mip-NeRF 360}} 
& 24.41 & 5840.30 & 26.61 & 0.796 & 0.245 & 16.43 & 4539.65 & 26.49 & 0.792 & 0.257 & \multirow{3}{*}{\textbf{-19.77\%}} & \multirow{3}{*}{\textbf{-4.30\%}} & \multirow{3}{*}{\textbf{+0.07 dB}} \\
& 18.05 & 4925.81 & 26.54 & 0.794 & 0.247 & 12.16 & 3890.42 & 26.28 & 0.788 & 0.263 & & & \\
& 11.38 & 3973.40 & 26.06 & 0.781 & 0.265 & 7.72 & 3395.97 & 25.56 & 0.768 & 0.289 & & & \\
\cmidrule{2-14}
\multirow{3}{*}{\textbf{Deep Blending}} 
& 23.57 & 5553.81 & 29.82 & 0.903 & 0.250 & 11.57 & 3492.71 & 29.82 & 0.906 & 0.258 & \multirow{3}{*}{\textbf{-30.92\%}} & \multirow{3}{*}{\textbf{-21.35\%}} & \multirow{3}{*}{\textbf{+0.25 dB}} \\
& 17.50 & 4664.81 & 29.82 & 0.903 & 0.254 & 8.58 & 3280.84 & 29.64 & 0.901 & 0.269 & & & \\
& 11.02 & 3642.29 & 29.54 & 0.896 & 0.269 & 5.50 & 2801.20 & 29.03 & 0.889 & 0.292 & & & \\
\cmidrule{2-14}
\multirow{3}{*}{\textbf{Tanks \& Temples}} 
& 12.56 & 4248.56 & 23.31 & 0.838 & 0.197 & 10.52 & 3419.86 & 23.36 & 0.836 & 0.203 & \multirow{3}{*}{\textbf{-19.81\%}} & \multirow{3}{*}{\textbf{-14.38\%}} & \multirow{3}{*}{\textbf{+0.13 dB}} \\
& 9.35 & 3673.83 & 23.19 & 0.830 & 0.209 & 7.83 & 2916.13 & 23.12 & 0.824 & 0.219 & & & \\
& 7.54 & 3164.61 & 22.81 & 0.814 & 0.224 & 5.12 & 2555.14 & 22.64 & 0.800 & 0.248 & & & \\
\cmidrule{2-14}
\multirow{3}{*}{\textbf{BungeeNeRF}} 
& 65.79 & 11658.14 & 27.18 & 0.901 & 0.118 & 36.57 & 7109.88 & 27.09 & 0.899 & 0.129 & \multirow{3}{*}{\textbf{-35.69\%}} & \multirow{3}{*}{N/A} & \multirow{3}{*}{N/A} \\
& 47.99 & 9018.83 & 27.35 & 0.903 & 0.119 & 26.67 & 5871.53 & 27.12 & 0.897 & 0.135 & & & \\
& 29.62 & 6374.47 & 27.21 & 0.894 & 0.135 & 16.53 & 4415.33 & 26.67 & 0.878 & 0.162 & & & \\
\midrule

% ================= Block: FCGS =================
\multicolumn{14}{@{}l}{\cellcolor{gray!20}\textbf{Compression Method: FCGS}~\cite{chen2025fast}} \\
\multirow{3}{*}{\textbf{Mip-NeRF 360}} 
& 41.68 & 386.81 & 27.35 & 0.804 & 0.233 & 34.08 & 181.79 & 27.28 & 0.802 & 0.239 & \multirow{3}{*}{\textbf{-77.72\%}} & \multirow{3}{*}{\textbf{-24.34\%}} & \multirow{3}{*}{\textbf{+0.17 dB}} \\
& 34.75 & 907.27 & 27.19 & 0.801 & 0.231 & 24.71 & 165.20 & 27.16 & 0.799 & 0.243 & & & \\
& 27.75 & 763.06 & 27.01 & 0.797 & 0.238 & 20.72 & 111.44 & 27.01 & 0.796 & 0.247 & & & \\
\cmidrule{2-14}
\multirow{3}{*}{\textbf{Deep Blending}} 
& 37.23 & 554.19 & 29.67 & 0.901 & 0.244 & 20.59 & 73.47 & 29.71 & 0.905 & 0.249 & \multirow{3}{*}{\textbf{-86.41\%}} & \multirow{3}{*}{\textbf{-53.99\%}} & \multirow{3}{*}{N/A} \\
& 27.62 & 675.78 & 29.49 & 0.899 & 0.247 & 14.75 & 62.63 & 29.60 & 0.903 & 0.251 & & & \\
& 24.31 & 224.61 & 29.24 & 0.896 & 0.252 & 12.56 & 61.55 & 29.47 & 0.902 & 0.254 & & & \\
\cmidrule{2-14}
\multirow{3}{*}{\textbf{Tanks \& Temples}} 
& 24.71 & 461.62 & 23.47 & 0.841 & 0.181 & 17.73 & 89.69 & 23.82 & 0.843 & 0.188 & \multirow{3}{*}{\textbf{-55.54\%}} & \multirow{3}{*}{N/A} & \multirow{3}{*}{\textbf{+0.47 dB}} \\
& 18.40 & 435.02 & 23.39 & 0.838 & 0.184 & 13.06 & 193.63 & 23.77 & 0.840 & 0.191 & & & \\
& 16.00 & 186.84 & 23.32 & 0.835 & 0.188 & 11.18 & 198.37 & 23.67 & 0.837 & 0.195 & & & \\
\cmidrule{2-14}
\multirow{1}{*}{\textbf{BungeeNeRF}} 
& - & - & - & - & - & - & - & - & - & - & \multirow{1}{*}{-} & \multirow{1}{*}{-} & \multirow{1}{*}{-}\\

\midrule

% ================= Block: GHAP =================
\multicolumn{14}{@{}l}{\cellcolor{gray!20}\textbf{Compression Method: GHAP}~\cite{wang2025gaussian}} \\
\multirow{3}{*}{\textbf{Mip-NeRF 360}} 
& 436.52 & 129.57 & 27.43 & 0.808 & 0.231 & 281.14 & 127.93 & 27.31 & 0.802 & 0.247 & \multirow{3}{*}{\textbf{-19.56\%}} & \multirow{3}{*}{\textbf{-9.29\%}} & \multirow{3}{*}{\textbf{+0.06 dB}} \\
& 311.69 & 121.37 & 27.29 & 0.802 & 0.245 & 200.73 & 86.35 & 27.10 & 0.794 & 0.263 & & & \\
& 186.95 & 112.02 & 27.00 & 0.790 & 0.269 & 120.39 & 77.70 & 26.73 & 0.778 & 0.291 & & & \\
\cmidrule{2-14}
\multirow{3}{*}{\textbf{Deep Blending}} 
& 410.42 & 108.53 & 29.81 & 0.903 & 0.245 & 186.02 & 64.26 & 29.75 & 0.905 & 0.251 & \multirow{3}{*}{\textbf{-27.84\%}} & \multirow{3}{*}{N/A} & \multirow{3}{*}{\textbf{-0.05 dB}} \\
& 293.08 & 94.18 & 29.82 & 0.903 & 0.249 & 132.83 & 76.99 & 29.70 & 0.904 & 0.256 & & & \\
& 175.83 & 91.61 & 29.79 & 0.902 & 0.257 & 79.67 & 71.14 & 29.66 & 0.903 & 0.264 & & & \\
\cmidrule{2-14}
\multirow{3}{*}{\textbf{Tanks \& Temples}} 
& 258.91 & 116.62 & 23.79 & 0.846 & 0.180 & 162.22 & 60.88 & 23.92 & 0.843 & 0.193 & \multirow{3}{*}{\textbf{-42.11\%}} & \multirow{3}{*}{\textbf{-39.07\%}} & \multirow{3}{*}{\textbf{+0.14 dB}} \\
& 184.86 & 92.72 & 23.76 & 0.842 & 0.192 & 115.82 & 49.98 & 23.80 & 0.836 & 0.207 & & & \\
& 110.87 & 52.65 & 23.67 & 0.832 & 0.212 & 69.46 & 40.80 & 23.64 & 0.823 & 0.231 & & & \\
\cmidrule{2-14}
\multirow{3}{*}{\textbf{BungeeNeRF}} 
& 1114.36 & 452.21 & 27.63 & 0.910 & 0.106 & 594.80 & 319.47 & 27.46 & 0.907 & 0.119 & \multirow{3}{*}{\textbf{-23.63\%}} & \multirow{3}{*}{\textbf{-26.10\%}} & \multirow{3}{*}{\textbf{+0.30 dB}} \\
& 795.87 & 316.60 & 27.41 & 0.903 & 0.119 & 424.80 & 251.97 & 27.12 & 0.897 & 0.138 & & & \\
& 477.50 & 260.45 & 26.94 & 0.889 & 0.146 & 254.86 & 214.57 & 26.47 & 0.876 & 0.174 & & & \\

\bottomrule
\end{tabular}
}
\end{table*}
\begin{table*}[htbp]
\centering
\caption{\textbf{Quantitative evaluation of our method in terms of 3DGS pruning.} Various pruning techniques are evaluated using the representations of vanilla 3DGS and SPARE-GS as inputs.}
\label{tab:pruning_perspective}
\setlength{\tabcolsep}{2.5pt} 
\renewcommand{\arraystretch}{0.85} 
\setlength{\aboverulesep}{2pt} 
\setlength{\belowrulesep}{2pt} 
\scriptsize 
\resizebox{\textwidth}{!}{
\begin{tabular}{@{}l ccccc ccccc ccc@{}}
\toprule
\multirow{2}{*}{\textbf{Dataset}} & \multicolumn{5}{c}{\textbf{Input: Vanilla 3DGS (Base)}} & \multicolumn{5}{c}{\textbf{Input: SPARE-GS (Ours)}} & \multicolumn{3}{c}{\textbf{Ours vs. Base}} \\
\cmidrule(lr){2-6} \cmidrule(lr){7-11} \cmidrule(l){12-14}
& \textbf{Size (MB)} & \textbf{Time (s)} & \textbf{PSNR} & \textbf{SSIM} & \textbf{LPIPS} & \textbf{Size (MB)} & \textbf{Time (s)} & \textbf{PSNR} & \textbf{SSIM} & \textbf{LPIPS} & \textbf{$\Delta$Time} $\downarrow$ & \textbf{BD-Rate} $\downarrow$ & \textbf{BD-PSNR} $\uparrow$ \\

\midrule
% ================= Block: PUP =================
\multicolumn{14}{@{}l}{\cellcolor{gray!20}\textbf{Pruning Method: PUP}~\cite{hanson2025pup}} \\
\multirow{3}{*}{\textbf{Mip-NeRF 360}} 
& 436.61 & 137.39 & 26.94 & 0.807 & 0.227 & 281.21 & 108.93 & 26.39 & 0.798 & 0.243 & \multirow{3}{*}{\textbf{-18.42\%}} & \multirow{3}{*}{\textbf{-22.18\%}} & \multirow{3}{*}{\textbf{+2.04 dB}} \\
& 311.87 & 132.38 & 24.66 & 0.781 & 0.248 & 200.87 & 108.40 & 23.05 & 0.753 & 0.275 & & & \\
& 187.12 & 130.16 & 20.62 & 0.703 & 0.305 & 120.52 & 108.92 & 18.87 & 0.651 & 0.343 & & & \\
\cmidrule{2-14}
\multirow{3}{*}{\textbf{Deep Blending}} 
& 410.46 & 110.27 & 29.77 & 0.906 & 0.239 & 186.05 & 73.00 & 29.61 & 0.906 & 0.248 & \multirow{3}{*}{\textbf{-33.00\%}} & \multirow{3}{*}{\textbf{-39.48\%}} & \multirow{3}{*}{\textbf{+3.57 dB}} \\
& 293.19 & 107.79 & 29.08 & 0.898 & 0.251 & 132.90 & 72.25 & 27.65 & 0.884 & 0.275 & & & \\
& 175.91 & 105.55 & 25.67 & 0.860 & 0.296 & 79.74 & 71.58 & 21.73 & 0.812 & 0.344 & & & \\
\cmidrule{2-14}
\multirow{3}{*}{\textbf{Tanks \& Temples}} 
& 258.98 & 65.04 & 23.06 & 0.840 & 0.182 & 162.26 & 49.02 & 22.96 & 0.834 & 0.196 & \multirow{3}{*}{\textbf{-24.06\%}} & \multirow{3}{*}{\textbf{-26.16\%}} & \multirow{3}{*}{\textbf{+2.03 dB}} \\
& 184.98 & 67.11 & 21.38 & 0.813 & 0.207 & 115.90 & 50.01 & 20.24 & 0.795 & 0.231 & & & \\
& 110.99 & 62.05 & 18.06 & 0.742 & 0.265 & 69.54 & 48.44 & 16.54 & 0.708 & 0.295 & & & \\
\cmidrule{2-14}
\multirow{3}{*}{\textbf{BungeeNeRF}} 
& 1114.41 & 212.16 & 27.68 & 0.912 & 0.099 & 594.84 & 109.42 & 27.38 & 0.907 & 0.112 & \multirow{3}{*}{\textbf{-45.84\%}} & \multirow{3}{*}{\textbf{-35.97\%}} & \multirow{3}{*}{\textbf{+2.65 dB}} \\
& 796.01 & 186.75 & 26.75 & 0.894 & 0.114 & 424.89 & 99.66 & 25.98 & 0.879 & 0.136 & & & \\
& 477.61 & 164.91 & 23.49 & 0.831 & 0.166 & 254.93 & 96.29 & 21.67 & 0.792 & 0.206 & & & \\

\midrule
% ================= Block: REFINE =================
\multicolumn{14}{@{}l}{\cellcolor{gray!20}\textbf{Pruning Method: REFINE}~\cite{chen2026refine}} \\
\multirow{3}{*}{\textbf{Mip-NeRF 360}} 
& 436.61 & 5.21 & 27.25 & 0.807 & 0.227 & 281.06 & 3.46 & 26.65 & 0.798 & 0.244 & \multirow{3}{*}{\textbf{-32.63\%}} & \multirow{3}{*}{\textbf{-18.55\%}} & \multirow{3}{*}{\textbf{+1.07 dB}} \\
& 311.87 & 4.44 & 26.21 & 0.789 & 0.247 & 200.91 & 3.73 & 25.20 & 0.770 & 0.272 & & & \\
& 187.12 & 4.48 & 23.55 & 0.733 & 0.299 & 120.57 & 2.33 & 21.89 & 0.680 & 0.340 & & & \\
\cmidrule{2-14}
\multirow{3}{*}{\textbf{Deep Blending}} 
& 410.46 & 3.79 & 29.59 & 0.900 & 0.245 & 186.05 & 2.52 & 29.40 & 0.900 & 0.255 & \multirow{3}{*}{\textbf{-43.08\%}} & \multirow{3}{*}{\textbf{-40.77\%}} & \multirow{3}{*}{\textbf{+1.59 dB}} \\
& 293.19 & 3.71 & 29.09 & 0.893 & 0.255 & 132.90 & 1.83 & 28.44 & 0.886 & 0.274 & & & \\
& 175.91 & 2.90 & 27.64 & 0.870 & 0.283 & 79.74 & 1.57 & 26.29 & 0.849 & 0.315 & & & \\
\cmidrule{2-14}
\multirow{3}{*}{\textbf{Tanks \& Temples}} 
& 258.98 & 3.24 & 23.35 & 0.839 & 0.182 & 162.26 & 1.66 & 23.27 & 0.835 & 0.194 & \multirow{3}{*}{\textbf{-33.11\%}} & \multirow{3}{*}{\textbf{-28.60\%}} & \multirow{3}{*}{\textbf{+1.43 dB}} \\
& 184.98 & 2.59 & 22.35 & 0.816 & 0.203 & 115.90 & 2.00 & 21.83 & 0.804 & 0.222 & & & \\
& 110.99 & 1.69 & 20.15 & 0.758 & 0.256 & 69.54 & 1.37 & 19.33 & 0.737 & 0.282 & & & \\
\cmidrule{2-14}
\multirow{3}{*}{\textbf{BungeeNeRF}} 
& 1114.41 & 14.20 & 26.95 & 0.904 & 0.109 & 594.84 & 6.24 & 26.45 & 0.893 & 0.127 & \multirow{3}{*}{\textbf{-49.08\%}} & \multirow{3}{*}{\textbf{-36.41\%}} & \multirow{3}{*}{\textbf{+2.63 dB}} \\
& 796.01 & 10.48 & 25.45 & 0.874 & 0.136 & 424.89 & 4.83 & 24.52 & 0.845 & 0.171 & & & \\
& 477.61 & 7.31 & 22.53 & 0.783 & 0.214 & 254.93 & 5.22 & 21.08 & 0.718 & 0.276 & & & \\

\midrule

% ================= Block: RAP =================
\multicolumn{14}{@{}l}{\cellcolor{gray!20}\textbf{Pruning Method: RAP}~\cite{yang2026rap}} \\
\multirow{3}{*}{\textbf{Mip-NeRF 360}} 
& 436.61 & 78.81 & 27.37 & 0.810 & 0.230 & 281.21 & 47.62 & 27.02 & 0.800 & 0.250 & \multirow{3}{*}{\textbf{-34.05\%}} & \multirow{3}{*}{\textbf{-21.54\%}} & \multirow{3}{*}{\textbf{+1.08 dB}} \\
& 311.87 & 74.15 & 26.47 & 0.790 & 0.250 & 200.87 & 59.70 & 25.70 & 0.770 & 0.280 & & & \\
& 187.12 & 71.81 & 24.08 & 0.730 & 0.304 & 120.52 & 40.91 & 23.12 & 0.697 & 0.342 & & & \\
\cmidrule{2-14}
\multirow{3}{*}{\textbf{Deep Blending}} 
& 410.46 & 79.44 & 29.72 & 0.905 & 0.243 & 186.05 & 30.97 & 29.48 & 0.904 & 0.254 & \multirow{3}{*}{\textbf{-57.70\%}} & \multirow{3}{*}{\textbf{-39.69\%}} & \multirow{3}{*}{\textbf{+1.46 dB}} \\
& 293.19 & 73.17 & 29.22 & 0.898 & 0.256 & 132.90 & 37.25 & 28.54 & 0.889 & 0.277 & & & \\
& 175.91 & 70.76 & 27.86 & 0.877 & 0.287 & 79.74 & 26.27 & 26.69 & 0.857 & 0.320 & & & \\
\cmidrule{2-14}
\multirow{3}{*}{\textbf{Tanks \& Temples}} 
& 258.98 & 42.74 & 23.61 & 0.847 & 0.176 & 162.26 & 25.09 & 23.63 & 0.840 & 0.191 & \multirow{3}{*}{\textbf{-37.22\%}} & \multirow{3}{*}{\textbf{-29.98\%}} & \multirow{3}{*}{\textbf{+1.16 dB}} \\
& 184.98 & 41.19 & 22.93 & 0.825 & 0.198 & 115.90 & 30.19 & 22.62 & 0.808 & 0.223 & & & \\
& 110.99 & 39.10 & 21.18 & 0.765 & 0.254 & 69.54 & 21.96 & 20.60 & 0.734 & 0.291 & & & \\
\cmidrule{2-14}
\multirow{1}{*}{\textbf{BungeeNeRF}} 
& - & - & - & - & - & - & - & - & - & - & \multirow{1}{*}{{-}} & \multirow{1}{*}{{-}} & \multirow{1}{*}{{-}} \\
\bottomrule
\end{tabular}
}
\end{table*}

In this section, we evaluated the effectiveness of SPARE-GS for downstream post-processing tasks. We first tested post-training compression, a task aimed at minimizing the storage footprint of 3D scenes. For this purpose, we evaluated five representative post-training compression frameworks: LightGaussian~\cite{fan2024lightgaussian}, SOG~\cite{morgenstern2024compact}, MesonGS~\cite{xie2024mesongs}, FCGS~\cite{chen2025fast}, and GHAP~\cite{wang2025gaussian}. Subsequently, we examined post-training pruning, which typically relies on specific criteria to eliminate primitives for further memory reduction. we evaluated three pruning techniques (PUP~\cite{hanson2025pup}, REFINE~\cite{chen2026refine}, and RAP~\cite{yang2026rap}) without fine-tuning. All post-processing frameworks take the 3D Gaussian representations optimized by the original 3DGS pipeline with our SPARE-GS integrated as their inputs.

\vspace{0.5em}
\noindent\textbf{Performance on 3DGS Compression.}
Table~\ref{tab:compression_perspective} shows that using SPARE-GS representations generally reduces both the resulting model size and post-processing time. For LightGaussian and SOG, the BD-Rate reductions range from 27.78\% to 55.38\% and from 35.60\% to 55.99\%, respectively. MesonGS exhibits more moderate but consistent gains on the datasets where BD-Rate can be computed, with reductions ranging from 4.30\% to 21.35\%, accompanied by processing-time reductions ranging from 19.77\% to 35.69\%. FCGS achieves BD-Rate reductions of 24.34\% and 53.99\% on Mip-NeRF 360 and Deep Blending, respectively, together with a BD-PSNR improvement of 0.47 dB on Tanks \& Temples. GHAP also benefits from the compact input representations, although the magnitude of its rate-distortion improvement varies across datasets. Specifically, it achieves BD-Rate reductions ranging from 9.29\% to 39.07\% where the metric is available, while its BD-PSNR changes range from -0.05 dB to +0.30 dB.

The processing-time results follow a similar trend. SOG reduces latency by up to 46.77\%, while FCGS achieves a reduction of up to 86.41\%. These improvements can be attributed to the smaller number of input primitives and the correspondingly reduced processing workload for downstream clustering, quantization, and coding operations.

\vspace{0.5em}
\noindent\textbf{Performance on 3DGS Pruning.}
As shown in Table~\ref{tab:pruning_perspective}, PUP achieves BD-Rate reductions ranging from 22.18\% to 39.48\%, BD-PSNR improvements ranging from 2.03 dB to 3.57 dB, and processing-time reductions ranging from 18.42\% to 45.84\%. REFINE obtains BD-Rate reductions ranging from 18.55\% to 40.77\%, together with BD-PSNR improvements ranging from 1.07 dB to 2.63 dB. RAP similarly achieves BD-Rate reductions ranging from 21.54\% to 39.69\%, BD-PSNR improvements ranging from 1.08 dB to 1.46 dB, and processing-time reductions ranging from 34.05\% to 57.70\%. These consistent trends across different pruning strategies suggest that regulating structural growth during training provides a more compact input representation for subsequent post-training simplification.

\vspace{0.5em}
\noindent\textbf{Summary.}
Across both compression and pruning evaluations, SPARE-GS generally reduces post-processing time and improves downstream rate-distortion efficiency. Although individual operating points may exhibit modest variations in PSNR, SSIM, or LPIPS, the reported BD metrics indicate more favorable quality-size trade-offs for most combinations of downstream methods and datasets where sufficient curve overlap is available. The magnitude of the improvement varies across downstream methods, but the overall results support SPARE-GS as an effective pre-conditioning strategy for subsequent 3DGS post-processing.

\subsection{Ablation Study} \label{sec:ablation}
In this section, we conducted ablation studies on the Mip-NeRF 360 dataset to evaluate the individual contributions of the proposed modules and analyze the sensitivity of the main hyperparameters. To ensure a fair comparison, all ablation variants, including the complete model, were re-evaluated under the same experimental settings.

\subsubsection{Effectiveness of Proposed Modules} 
\begin{figure}[]
\centering
% ==================== 1. 表格部分 ====================
\captionof{table}{\textbf{Ablation Study on Core Components.} The individual contributions of Budget-modulated Densification (\textit{BMD}), Redundancy-aware Pruning (\textit{RAP}), and Equilibrium-driven Termination (\textit{EDT}) are evaluated by removing each module from the complete pipeline.}
\label{tab:ablation_modules}
\setlength{\tabcolsep}{2pt} 
\footnotesize 
\resizebox{\linewidth}{!}{
\begin{tabular}{@{}l ccc ccccc@{}}
\toprule
\textbf{Method} & \textit{BMD} & \textit{RAP} & \textit{EDT} & \textbf{Gaussians}$\downarrow$ & \textbf{Time(s)}$\downarrow$ & \textbf{PSNR}$\uparrow$ & \textbf{SSIM}$\uparrow$ & \textbf{LPIPS}$\downarrow$ \\
\midrule
Baseline & \texttimes & \texttimes & \texttimes & 2.64M & 1314 & {27.51} & {0.813} & {0.221} \\
\textit{w/o} \textit{BMD} & \texttimes & \checkmark & \checkmark & 1.88M & 1091 & 27.50 & 0.813 & 0.227 \\
\textit{w/o} \textit{RAP} & \checkmark & \texttimes & \checkmark & 2.32M & 1161 & 27.48 & 0.809 & 0.228 \\
\textit{w/o} \textit{EDT} & \checkmark & \checkmark & \texttimes & 1.68M & 964 & 27.45 & 0.809 & 0.233 \\
\midrule
\textbf{Full} & \checkmark & \checkmark & \checkmark & {1.69M} & {928} & 27.46 & 0.809 & 0.234 \\
\bottomrule
\end{tabular}
}
    
    % ==================== 分割间距 ====================
    \vspace{2em} 
    
    % ==================== 2. 图片部分 ====================
    \subfloat[bicycle]{
        \includegraphics[width=0.22\textwidth]{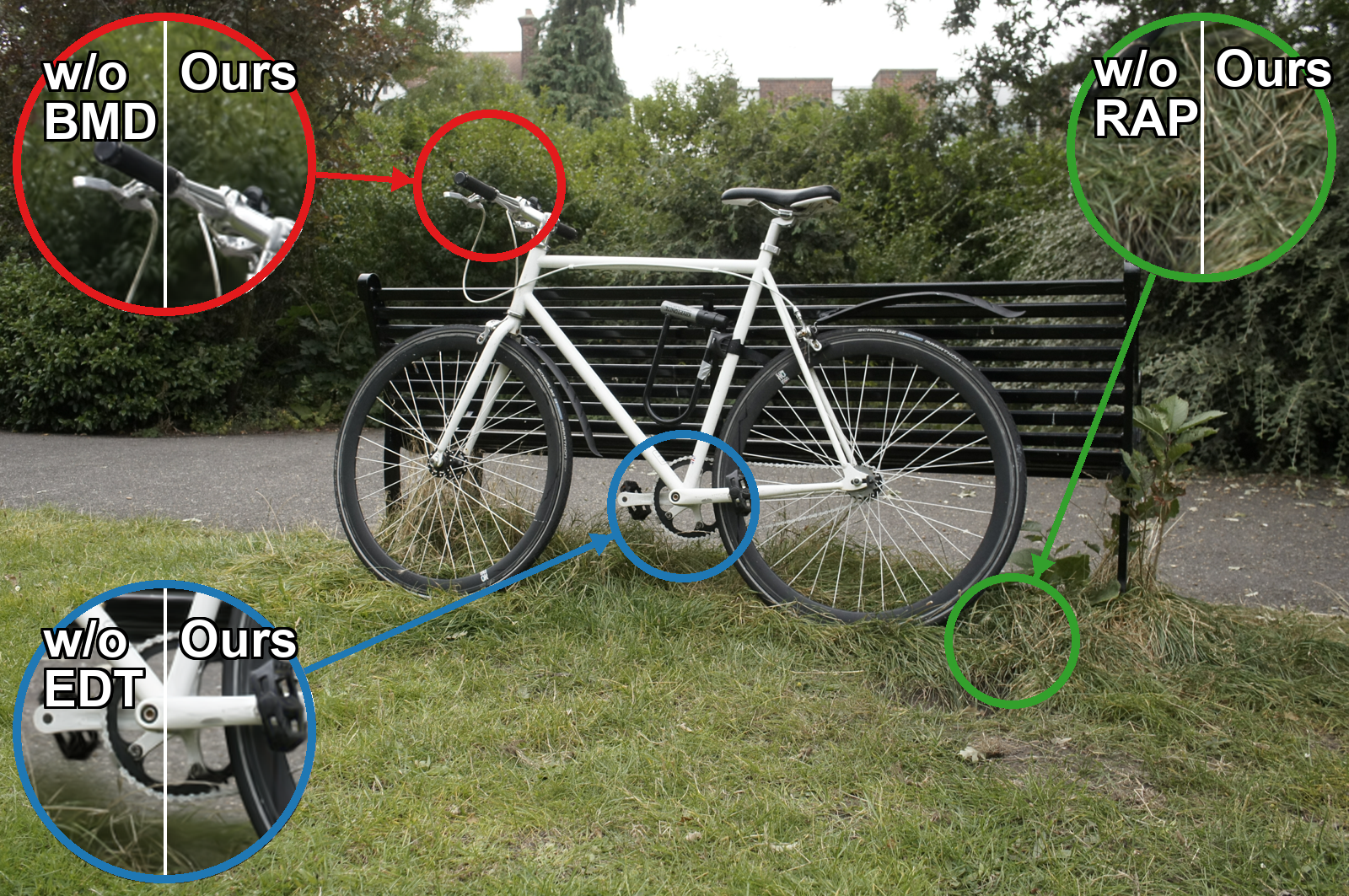}
    }\hfill 
    \subfloat[bonsai]{
        \includegraphics[width=0.22\textwidth]{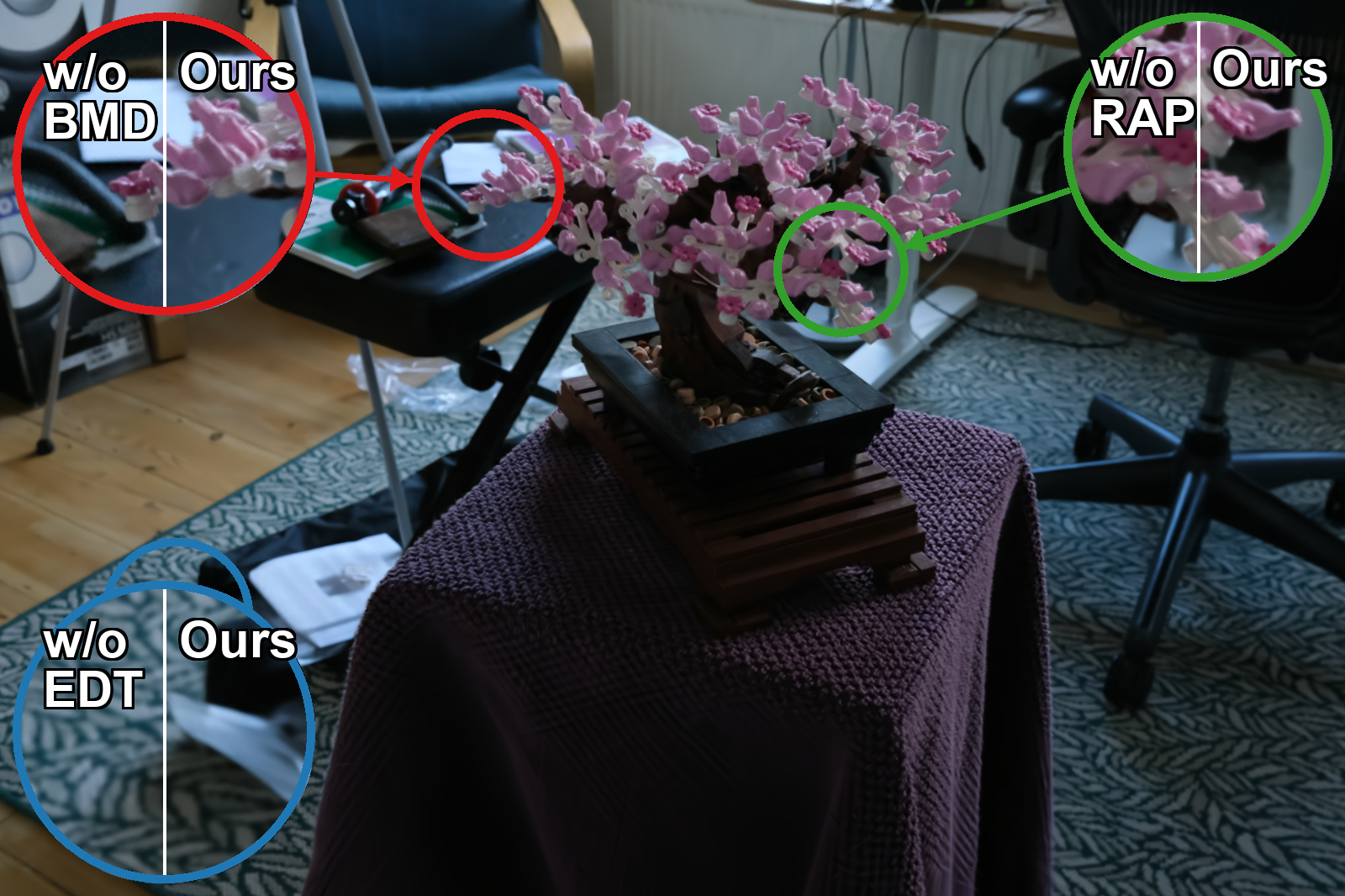}
    }
    \captionof{figure}{\textbf{Visual comparison of ablation study.} Zooming in for details.}
    \label{fig:addtionvisual2}
\end{figure}

As detailed in Sec.~\ref{sec4.3}, SPARE-GS comprises three components: Budget-modulated Densification, \textit{BMD}, Redundancy-aware Pruning, \textit{RAP}, and Equilibrium-driven Termination, \textit{EDT}. We conducted a leave-one-out ablation study by removing one component at a time and comparing each resulting variant with both the baseline and the complete model. The results are summarized in Table~\ref{tab:ablation_modules}.

\vspace{0.5em}
\noindent\textbf{Impact of Budget-modulated Densification.} 
\textit{BMD} acts as the stage regulator of structural growth. Instead of allowing the base densification rule to expand all high-gradient primitives uniformly, it reweights the densification score according to the regional budget deviation. Removing \textit{BMD} increases the primitive count from 1.69M to 1.88M and the training time from 928.40 s to 1090.97 s. This corresponds to around 11.1\% more Gaussians and 17.5\% longer training compared with the full model. The larger increase in training time suggests that unconstrained densification not only leaves more primitives in the final representation but also introduces additional optimization overhead throughout subsequent iterations. Although \textit{w/o} \textit{BMD} achieves a slightly higher PSNR, this improvement is only 0.04 dB over the full model and comes with a clear increase in structural costs. This confirms that \textit{BMD} provides an effective front-end control mechanism that suppresses unnecessary growth while preserving comparable rendering quality.

\vspace{0.5em}
\noindent\textbf{Impact of Redundancy-aware Pruning.} 
\textit{RAP} serves as the complementary back-end regulator to \textit{BMD}. While \textit{BMD} suppresses unnecessary growth before new primitives are created, \textit{RAP} removes redundant primitives that have already accumulated in over-budgeted regions. This role is clearly reflected by the \textit{w/o} \textit{RAP} variant, where the number of Gaussians increases to 2.32M, which is 36.8\% higher than the full model. The training time also rises from 928.40 s to 1160.72 s, indicating that redundant primitives not only enlarge the final representation but also continuously increase rendering and optimization costs during training. Meanwhile, the rendering quality remains nearly unchanged, with only a 0.02 dB PSNR difference between \textit{w/o} \textit{RAP} and the full model. This suggests that most primitives removed by \textit{RAP} are indeed low-contribution structural redundancies rather than essential high-frequency details. Therefore, \textit{RAP} is critical for converting the budget feedback signal into actual model compactness.

\vspace{0.5em}
\noindent\textbf{Impact of Equilibrium-driven Termination.}
\textit{EDT} mainly targets the temporal redundancy of training rather than directly reducing the final primitive count. Compared with the full model, the \textit{w/o} \textit{EDT} variant continues optimization for a longer time, increasing the training cost from 928.40 s to 964.34 s. However, this additional optimization does not translate into meaningful quality improvements: PSNR changes from 27.46 dB to 27.45 dB, while SSIM remains the same and LPIPS differs only marginally. The primitive count of \textit{w/o} \textit{EDT} is also very close to that of the full model, indicating that the structural state has already become saturated when \textit{EDT} triggers termination. Therefore, \textit{EDT} effectively identifies the low-return stage of training, where continuing the fixed schedule mostly incurs computational costs without providing noticeable reconstruction benefits.

\vspace{0.5em}
Overall, the ablation results reveal a clear division of labor among the three modules. \textit{BMD} prevents excessive growth by regulating densification before new primitives are introduced, \textit{RAP} removes low-contribution primitives accumulated in over-budgeted regions, and \textit{EDT} avoids redundant optimization after the structural state becomes saturated. Their combination yields a favorable efficiency-quality trade-off: compared with the baseline, the full model removes nearly one million Gaussians and reduces training time by 29.3\%, while maintaining comparable rendering fidelity. 

\vspace{0.5em}
\subsubsection{Hyperparameter Sensitivity Analysis}
We further analyzed the sensitivity of representative hyperparameters introduced in our framework, including the spatial grid resolution, the pruning opacity threshold, and the thresholds governing equilibrium-driven termination. The quantitative results are presented in Table~\ref{tab:ablation_hyper}.

\begin{table}[htbp] % 加上 htbp 增强表格浮动排版的灵活性
\centering
\caption{\textbf{Sensitivity analysis of key hyperparameters.} Default settings used in our full model are marked in \colorbox{gray!20}{gray}.}
\label{tab:ablation_hyper}
\setlength{\tabcolsep}{3pt} 
\resizebox{\linewidth}{!}{
\begin{tabular}{@{}l ccccc@{}}
\toprule
\textbf{Hyperparameters} & \textbf{Gaussians} $\downarrow$ & \textbf{Time (s)} $\downarrow$ & \textbf{PSNR} $\uparrow$ & \textbf{SSIM} $\uparrow$ & \textbf{LPIPS} $\downarrow$ \\

\midrule
\multicolumn{6}{@{}l}{\textit{$K$}} \\
$16^3$ & \textbf{1,685,685} & \textbf{888.92} & 27.43 & \textbf{0.809} & \textbf{0.234} \\
\rowcolor{gray!20} $32^3$ (Default) & 1,693,989 & 928.40 & \textbf{27.46} & \textbf{0.809} & \textbf{0.234} \\
$64^3$ & 1,688,937 & 894.68 & 27.42 & \textbf{0.809} & \textbf{0.234} \\

\midrule
\multicolumn{6}{@{}l}{\textit{$\tau_\alpha$}} \\
$0.01$ & 2,230,887 & 1136.96 & \textbf{27.48} & \textbf{0.809} & \textbf{0.228} \\
\rowcolor{gray!20} $0.05$ (Default) & 1,693,989 & 928.40 & 27.46 & \textbf{0.809} & 0.234 \\
$0.10$ & \textbf{1,313,693} & \textbf{744.71} & 27.34 & 0.804 & 0.245 \\

\midrule
\multicolumn{6}{@{}l}{\textit{$\ell$}} \\
$1$ step & \textbf{1,688,295} & \textbf{788.26} & 27.31 & 0.807 & 0.236 \\
\rowcolor{gray!20} $2$ steps (Default) & 1,693,989 & 928.40 & \textbf{27.46} & \textbf{0.809} & \textbf{0.234} \\
$3$ steps & 1,694,566 & 937.14 & \textbf{27.46} & \textbf{0.809} & \textbf{0.234} \\

\midrule
\multicolumn{6}{@{}l}{\textit{$\epsilon_L$}} \\
$0.016$ & \textbf{1,692,243} & \textbf{868.78} & 27.43 & 0.808 & 0.235 \\
\rowcolor{gray!20} $0.008$ (Default) & 1,693,989 & 928.40 & 27.46 & \textbf{0.809} & \textbf{0.234} \\
$0.004$ & 1,692,740 & 962.89 & \textbf{27.49} & \textbf{0.809} & \textbf{0.234} \\

\midrule
\multicolumn{6}{@{}l}{\textit{$\epsilon_N$}} \\
$0.05$ & 1,694,729 & \textbf{898.40} & 27.37 & 0.808 & \textbf{0.234} \\
\rowcolor{gray!20} $0.01$ (Default) & \textbf{1,693,989} & 928.40 & \textbf{27.46} & \textbf{0.809} & \textbf{0.234} \\
$0.001$ & 1,694,289 & 941.40 & \textbf{27.46} & \textbf{0.809} & \textbf{0.234} \\

\bottomrule
\end{tabular}
}
\end{table}

\vspace{0.5em}
\noindent\textbf{Spatial Grid Size $K$.} 
The spatial grid resolution controls the granularity at which regional statistics are aggregated and target budgets are assigned. The three resolutions produce similar Gaussian counts and perceptual metrics, indicating limited sensitivity to the grid granularity. Although $16^3$ is slightly faster, $32^3$ achieves the highest PSNR while providing finer spatial localization than the coarse grid. We therefore adopt $32^3$ as a conservative middle-resolution setting.

\vspace{0.5em}
\noindent\textbf{Pruning Opacity Threshold $\tau_\alpha$.} 
This parameter, defined in Eq.~\eqref{eq:budget_prune}, controls the aggressiveness of redundancy removal. Setting a high threshold of $\tau_\alpha = 0.10$ aggressively removes Gaussians, achieving a highly compact model of 1.31 M primitives. However, this leads to a noticeable degradation in rendering quality, where PSNR drops to 27.34 dB and LPIPS spikes to 0.245, indicating lost perceptual details. Conversely, a very conservative threshold of $\tau_\alpha = 0.01$ retains 2.23 M primitives, which achieves a better LPIPS of 0.228 at the cost of limiting the compression capability. Therefore, setting $\tau_\alpha = 0.05$ strikes the balance between representational capacity and structural parsimony.

\vspace{0.5em}
\noindent\textbf{Termination Patience $\ell$.} 
This parameter dictates the number of consecutive intervals required to trigger the Equilibrium-driven Termination established in Eq.~\eqref{eq:delta_loss_num}. A minimal patience of $\ell=1$ causes the optimization to halt prematurely before full convergence, degrading the final rendering fidelity with PSNR dropping to 27.31 dB and LPIPS increasing to 0.236. Our default setting of $\ell=2$ mitigates such premature termination and serves as a stable indicator of structural saturation.

\vspace{0.5em}
\noindent\textbf{Loss Saturation Threshold $\epsilon_L$.} 
The loss tolerance $\epsilon_L$ determines the sensitivity of the photometric convergence check. A loose tolerance of $\epsilon_L=0.016$ can trigger the halting condition prematurely, leading to sub-optimal rendering quality. Alternatively, setting a tight tolerance of $\epsilon_L=0.004$ prolongs the training to 962 s without yielding meaningful perceptual quality gains over the default configuration. We set $\epsilon_L=0.008$ to maintain a practical balance between training efficiency and convergence quality.

\vspace{0.5em}
\noindent\textbf{Count Saturation Threshold $\epsilon_N$.} 
We additionally evaluate the threshold for primitive count variation, $\epsilon_N$, which acts as an explicit indicator of structural convergence. A loose count threshold of $\epsilon_N=0.05$ halts the structural evolution before full optimization, reducing the PSNR to 27.37 dB. Conversely, a strict threshold of $\epsilon_N=0.001$ fails to detect saturation in time, extending the training duration to 941 s without improving rendering quality or LPIPS. Our default setting of $\epsilon_N=0.01$ preserves photometric accuracy while preventing redundant computations.

\vspace{0.5em}
Overall, the default configuration consistently provides a balanced trade-off between compactness, training efficiency, and rendering fidelity, while more aggressive settings mainly trade quality for additional compression or speed.

\subsection{Further Analysis}
To examine the effectiveness of the marginal utility proxy, we conducted a leave-one-out ablation study by removing the regional gradient, multi-view visibility, or geometric capacity normalization while retaining the remaining components. 

\begin{table}[htbp]
\centering
\caption{\textbf{Ablation Study on the Marginal Utility Proxy.} }
\label{tab:ablation_proxy}
\setlength{\tabcolsep}{2.5pt}
\footnotesize
\resizebox{\linewidth}{!}{
\begin{tabular}{@{}l ccc ccccc@{}}
\toprule
\textbf{Method} & ${H^{(t)}_r}$ & ${V^{(t)}_r}$ & ${C^{(t)}_r}$ & \textbf{Gaussians}$\downarrow$ & \textbf{Time(s)}$\downarrow$ & \textbf{PSNR}$\uparrow$ & \textbf{SSIM}$\uparrow$ & \textbf{LPIPS}$\downarrow$ \\
\midrule
\textit{w/o Gradient} & \texttimes & \checkmark & \checkmark & 1.72M & 972 & 27.45 & 0.809 & 0.234 \\
\textit{w/o Visibility} & \checkmark & \texttimes & \checkmark & 1.73M & 930 & 27.46 & 0.808 & 0.235 \\
\textit{w/o Capacity} & \checkmark & \checkmark & \texttimes & 1.73M & 961 & 27.46 & 0.808 & 0.234 \\
\midrule
\textbf{Full} & \checkmark & \checkmark & \checkmark & 1.69M & 928 & 27.46 & 0.809 & 0.234 \\
\bottomrule
\end{tabular}
}
\end{table}

As shown in Table~\ref{tab:ablation_proxy}, removing any component increases the final Gaussian count while preserving comparable rendering quality. Gradient magnitude and capacity normalization have clearer effects on training efficiency, increasing the training time by 4.67\% and 3.56\%, respectively, when removed. Multi-view visibility mainly contributes to structural compactness, with its removal increasing the Gaussian count by 1.90\%, although its effect on training time is limited. These results indicate that the three signals provide complementary, rather than individually dominant, guidance for regional allocation.

\begin{figure}[]
    \centering
    \includegraphics[width=0.85\linewidth]{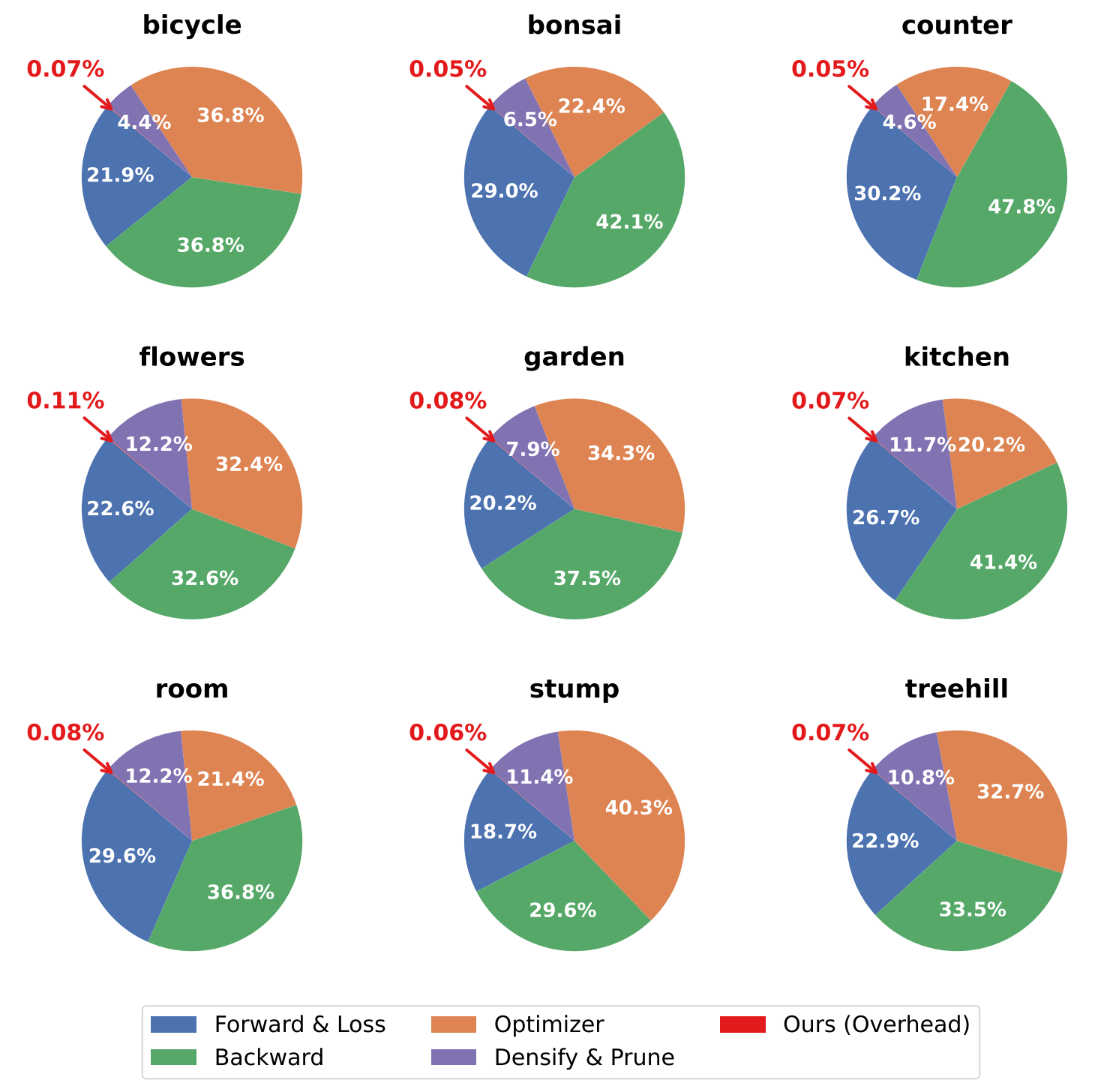}
    \caption{\textbf{Computational time breakdown.} The pie charts illustrate the temporal distribution of different training components. The overhead introduced by SPARE-GS is highlighted in red. }
    \label{fig:overhead_pie}
\end{figure}

\vspace{0.5em}
\noindent\textbf{Computational Time Breakdown.} 
To verify the computational efficiency of SPARE-GS, we profiled the runtime breakdown across the nine Mip-NeRF 360 scenes. As illustrated in Fig.~\ref{fig:overhead_pie}, the overhead introduced by our online budget module is practically negligible. For each scene, the accumulated budget-estimation time ranges from 0.33 s to 0.93 s over the entire training process. This constitutes a mere 0.073\% of the total training time on average, occupying just 0.87\% of the structural update phase. In contrast, standard operations (i.e., backward propagation, optimizer updates, and forward rendering) dominate the runtime. 
This near-zero overhead stems from two design choices. First, SPARE-GS directly reuses statistics natively maintained during standard 3DGS training (e.g., accumulated gradients, visibility, and opacity), avoiding redundant metric computations. Second, our lightweight operations, including voxel assignment, regional aggregation, and budget allocation, are executed exclusively at structural update intervals rather than at every rendering step. Consequently, the acceleration reported in Table~\ref{tab:comprehensive_results} primarily stems from the reduced structural redundancies and adaptive termination.

\begin{table}[]
\centering
\caption{\textbf{Representative Failure cases on Mip-NeRF 360.}}
\label{tab:failure_cases}
\setlength{\tabcolsep}{2.5pt} 
\renewcommand{\arraystretch}{1.1} 
\begin{tabular}{@{}l cc ccc @{}}
\toprule
\multirow{2}{*}{\textbf{Scene}} & \multicolumn{2}{c}{\textbf{Efficiency gains} ($\downarrow$)} & \multicolumn{3}{c}{\textbf{Quality degradation}} \\
\cmidrule(lr){2-3} \cmidrule(l){4-6}
& \textbf{$\Delta$Gaussians} & \textbf{$\Delta$Time} & \textbf{$\Delta$PSNR} & \textbf{$\Delta$SSIM} & \textbf{$\Delta$LPIPS} \\
\midrule
bonsai  & $-$33.71\% & $-$19.13\% & $-$0.28 & $-$0.002 & $+$0.007 \\
flowers & $-$32.73\% & $-$24.97\% & $-$0.20 & $-$0.009 & $+$0.016 \\
garden  & $-$33.24\% & $-$44.77\% & $-$0.24 & $-$0.008 & $+$0.016 \\
\bottomrule
\end{tabular}
\end{table}
\vspace{0.5em}

\vspace{0.5em}
\noindent\textbf{Analysis of Structural Regulation.} 
To understand how SPARE-GS regulates structural growth, we analyzed the active densification phase across the Mip-NeRF 360 scenes. As optimization progresses, SPARE-GS consistently suppresses excessive Gaussian proliferation. The average reduction in primitive count widens steadily from 8.7\% at 1.5K iterations to 26.5\% at 6K, ultimately reaching 35.8\% by the end of the densification stage (15K). Scene-wise reductions at 15K iterations remain highly consistent, ranging from 29.7\% to 48.0\%.
This steady divergence validates that SPARE-GS is not merely a post-hoc pruning filter for an already overgrown model. Instead, it proactively reshapes the evolutionary trajectory by continuously penalizing redundant growth in over-budgeted regions and guiding new primitives toward a more compact allocation. These results confirm that the proposed budget-guided regulation effectively controls Gaussian proliferation during the critical structural growth phase across diverse scenes.

\begin{figure}[]
    \centering
    \includegraphics[width=0.95\linewidth]{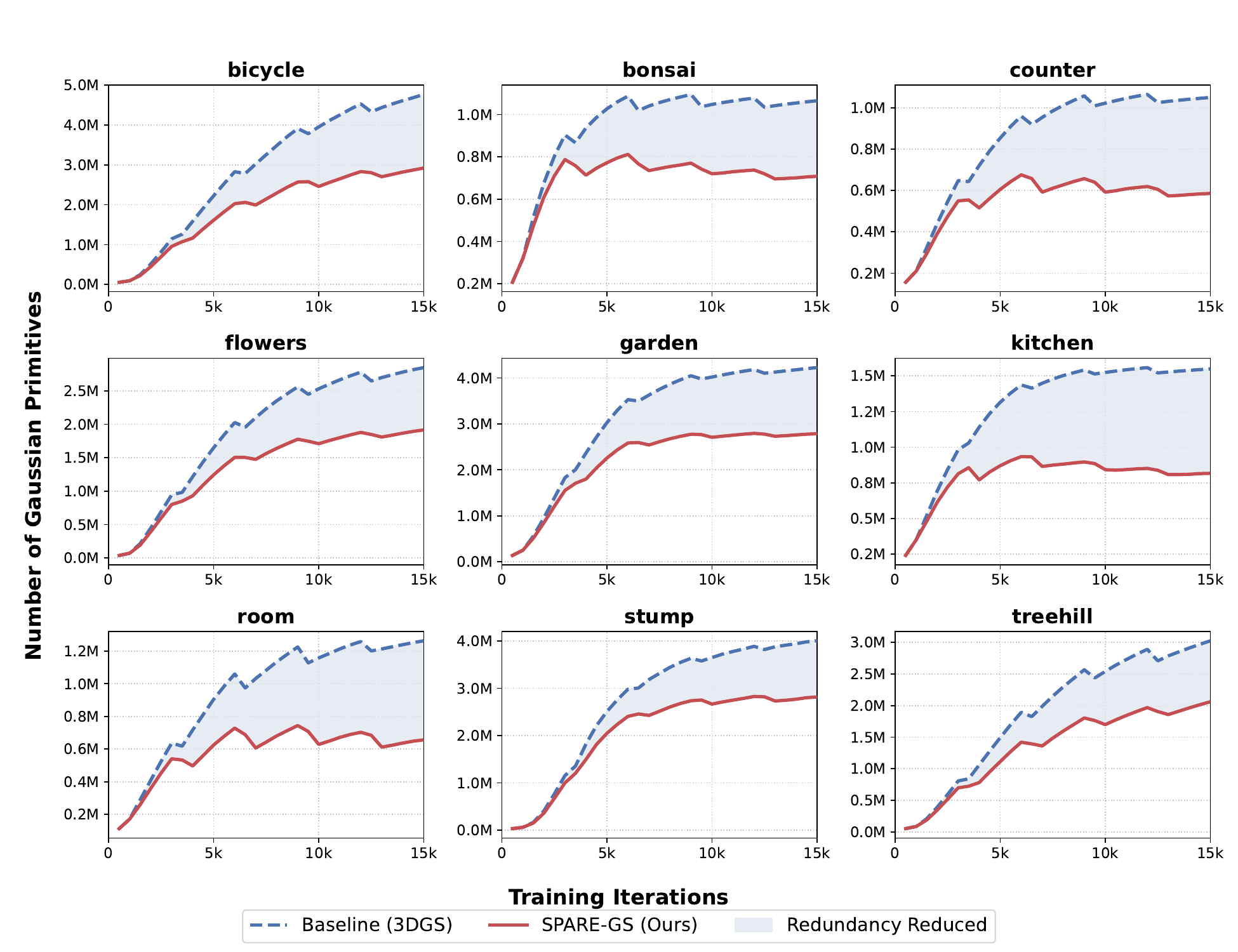}
    \caption{\textbf{Evolution of Gaussian count during the densification phase (0--15k iterations).} The shaded area represents the computational and memory footprint saved by our region-budget-aware regulation over time.}
    \label{fig:evolution_curves}
\end{figure}

\noindent\textbf{Failure Cases and Limitations.} \label{sec:failure_cases}
Table~\ref{tab:failure_cases} presents representative cases where the efficiency gains of SPARE-GS are accompanied by more noticeable quality degradation. Across the three scenes, SPARE-GS reduces the Gaussian count by approximately 33\% and the training time by 19.13\% to 44.77\%, while PSNR decreases by 0.20 dB to 0.28 dB. These cases may reflect limitations of the current heuristic demand estimation and capacity calibration in scenes with complex occlusions, fine details, or uneven view coverage. Future work could improve robustness by incorporating uncertainty-aware, occlusion-aware, or learned demand estimation, together with more conservative budget control in low-confidence regions.

\section{conclusion} \label{sec6}
In this work, we revisited the efficiency problem of 3DGS from the perspective of global structural resource allocation. Instead of treating primitive growth as a collection of independent local decisions, we formulated the structural evolution of 3DGS as a budget-constrained optimization problem and introduced SPARE-GS to dynamically align primitive allocation with regional representational demand. Through KKT-inspired budget assignment and feedback regulation, SPARE-GS provides a unified mechanism for coordinating structural growth, redundancy removal, and adaptive termination, thereby guiding the representation toward a more balanced efficiency-quality trade-off. Extensive experiments across standard standard, accelerated, and structure-enhanced 3DGS pipelines, as well as downstream compression and pruning methods, demonstrate the broad applicability of this formulation. Overall, SPARE-GS establishes a general framework for reasoning about and regulating finite structural resources in Gaussian-based scene representations. Future work will explore more accurate uncertainty-aware or learned demand estimation to improve robustness in structurally complex scenes.

%\medskip

{
\small % 对应官方要求：It is permissible to reduce the font size to \small
\bibliographystyle{IEEEtran} 
\bibliography{reference}

@String(PAMI  = {IEEE TPAMI})

@String(CVPR  = {CVPR})

@String(ECCV  = {ECCV})

@String(NeurIPS = {NeurIPS})

@String(ICML  = {ICML})

@String(ICLR  = {ICLR})

@String(BMVC  =	{BMVC})

@String(AAAI  = {AAAI})

@String(TOG   = {ACM TOG})

@String(TIP   = {IEEE TIP})

@String(TCSVT = {IEEE TCSVT})

@String(ACMMM = {ACM MM})

@String(PAMI  = {IEEE Trans. Pattern Anal. Mach. Intell.})

@String(CVPR  = {IEEE Conf. Comput. Vis. Pattern Recog.})

@String(ECCV  = {Eur. Conf. Comput. Vis.})

@String(NeurIPS = {Adv. Neural Inform. Process. Syst.})

@String(ICML  = {Int. Conf. Mach. Learn.})

@String(ICLR  = {Int. Conf. Learn. Represent.})

@String(BMVC  = {Brit. Mach. Vis. Conf.})

@String(TOG   = {ACM Trans. Graph.})

@String(TIP   = {IEEE Trans. Image Process.})

@String(TCSVT = {IEEE Trans. Circuit Syst. Video Technol.})

@String(ACMMM = {ACM Int. Conf. Multimedia})

@String(CGF   = {Comput. Graph. Forum})

@article{kerbl20233d,
  title   = {{3D} {Gaussian} splatting for real-time radiance field rendering},
  author  = {Kerbl, Bernhard and Kopanas, Georgios and Leimk{\"u}hler, Thomas and Drettakis, George},
  journal = TOG,
  volume  = {42},
  number  = {4},
  pages   = {139:1--139:14},
  year    = {2023}
}

@inproceedings{hanson2025pup,
  author    = {Hanson, Alex and Tu, Allen and Singla, Vasu and Jayawardhana, Mayuka and Zwicker, Matthias and Goldstein, Tom},
  title     = {{PUP} {3D-GS}: Principled uncertainty pruning for {3D} {Gaussian} splatting},
  booktitle = CVPR,
  pages     = {5949--5958},
  year      = {2025}
}

@inproceedings{fan2024lightgaussian,
  author    = {Fan, Zhiwen and Wang, Kevin and Wen, Kairun and Zhu, Zehao and Xu, Dejia and Wang, Zhangyang},
  title     = {{LightGaussian}: Unbounded {3D} {Gaussian} compression with 15x reduction and 200+ {FPS}},
  booktitle = NeurIPS,
  volume={37},
  pages={140138--140158},
  year={2024}
}

@inproceedings{xie2024mesongs,
  author    = {Xie, Shuzhao and Zhang, Weixiang and Tang, Chen and Bai, Yunpeng and Lu, Rongwei and Ge, Shijia and Wang, Zhi},
  title     = {{MesonGS}: Post-training compression of {3D} {Gaussians} via efficient attribute transformation},
  booktitle = ECCV,
  pages     = {434--452},
  year      = {2024}
}

@inproceedings{barron2022mipnerf360,
  author    = {Barron, Jonathan T and Mildenhall, Ben and Verbin, Dor and Srinivasan, Pratul P and Hedman, Peter},
  title     = {{Mip-NeRF} 360: Unbounded anti-aliased neural radiance fields},
  booktitle = CVPR,
  pages     = {5470--5479},
  year      = {2022}
}

@article{hedman2018deep,
  author  = {Hedman, Peter and Philip, Julien and Price, True and Frahm, Jan-Michael and Drettakis, George and Brostow, Gabriel},
  title   = {Deep blending for free-viewpoint image-based rendering},
  journal = TOG,
  volume  = {37},
  number  = {6},
  pages   = {1--15},
  year    = {2018}
}

@article{knapitsch2017tanks,
  author  = {Knapitsch, Arno and Park, Jaesik and Zhou, Qian-Yi and Koltun, Vladlen},
  title   = {{Tanks} and {Temples}: Benchmarking large-scale scene reconstruction},
  journal = TOG,
  volume  = {36},
  number  = {4},
  pages   = {1--13},
  year    = {2017}
}

@inproceedings{niemeyer2024radsplat,
  author    = {Niemeyer, Michael and Manhardt, Fabian and Rakotosaona, Marie-Julie and Oechsle, Michael and Duckworth, Daniel and Gosula, Rama and Tateno, Keisuke and Bates, John and Kaeser, Dominik and Tombari, Federico},
  title     = {{RadSplat}: Radiance field-informed {Gaussian} splatting for robust real-time rendering with 900+ {FPS}},
  booktitle = {International Conference on 3D Vision (3DV)},
  pages     = {134--144},
  year      = {2025}
}

@inproceedings{hanson2025speedy,
  title={Speedy-splat: Fast {3D} {Gaussian} Splatting with Sparse Pixels and Sparse Primitives},
  author={Hanson, Alex and Tu, Allen and Lin, Geng and Singla, Vasu and Zwicker, Matthias and Goldstein, Tom},
  booktitle = CVPR,
  pages={21537--21546},
  year={2025}
}

@inproceedings{chen2025fast,
  title={Fast Feedforward {3D} {Gaussian} Splatting Compression},
  author={Chen, Yihang and Wu, Qianyi and Li, Mengyao and Lin, Weiyao and Harandi, Mehrtash and Cai, Jianfei},
  booktitle = ICLR,
  volume={2025},
  pages={74859--74872},
  year={2025}
}

@inproceedings{chen2024hac,
  author    = {Yihang Chen and Qianyi Wu and Weiyao Lin and Mehrtash Harandi and Jianfei Cai},
  title     = {{HAC}: Hash-grid Assisted Context for {3D} {Gaussian} Splatting Compression},
  booktitle = ECCV,
  pages={422--438},
  year= {2024}
}

@inproceedings{ali2024trimming,
  author    = {Muhammad Salman Ali and Maryam Qamar and Sung Ho Bae and Enzo Tartaglione},
  title     = {Trimming the Fat: Efficient Compression of {3D} {Gaussian} Splats through Pruning},
  booktitle = BMVC,
  year      = {2024}
}

@inproceedings{chen2025haifgs,
  author    = {Jianing Chen and Zehao Li and Yujun Cai and Hao Jiang and Chengxuan Qian and Juyuan Kang and Shuqin Gao and Honglong Zhao and Tianlu Mao and Yucheng Zhang},
  title     = {{HAIF-GS}: Hierarchical and Induced Flow-Guided {Gaussian} Splatting for Dynamic Scene},
  booktitle = NeurIPS,
  volume={38},
  pages={125539--125563},
  year= {2025}
}

@inproceedings{wang2025gaussian,
author={Tao Wang and Mengyu Li and Geduo Zeng and Cheng Meng and Qiong Zhang},
title={Gaussian Herding across Pens: An Optimal Transport Perspective on Global Gaussian Reduction for 3{DGS}},
booktitle = NeurIPS,
volume={38},
pages={157898--157923},
year={2025}
}

@inproceedings{fang2024mini,
  author    = {Guangchi Fang and Bing Wang},
  title     = {Mini-Splatting: Representing Scenes with a Constrained Number of {Gaussians}},
  booktitle = ECCV,
  pages={165--181},
  year      = {2024}
}

@inproceedings{girish2024eagles,
  author    = {Sharath Girish and Kamal Gupta and Abhinav Shrivastava},
  title     = {Eagles: Efficient Accelerated {3D} {Gaussians} with Lightweight Encodings},
  booktitle = ECCV,
  pages={54--71},
  year      = {2024}
}

@inproceedings{lee2024compact,
  author    = {Lee, Joo Chan and Rho, Daniel and Sun, Xiangyu and Ko, Jong Hwan and Park, Eunbyung},
  title     = {Compact {3D} {Gaussian} Representation for Radiance Field},
  booktitle = CVPR,
  pages={21719--21728},
  year      = {2024}
}

@inproceedings{liu2024compgs,
  author    = {Xiangrui Liu and Xinju Wu and Pingping Zhang and Shiqi Wang and Zhu Li and Sam Kwong},
  title     = {{CompGS}: Efficient {3D} Scene Representation via Compressed {Gaussian} Splatting},
  booktitle = ACMMM,
  pages={2936--2944},
  year      = {2024}
}

@inproceedings{lu2024scaffoldgs,
  author    = {Tao Lu and Mulin Yu and Linning Xu and Yuanbo Xiangli and Limin Wang and Dahua Lin and Bo Dai},
  title     = {{Scaffold-GS}: Structured {3D} {Gaussians} for View-Adaptive Rendering},
  booktitle = CVPR,
  pages={20654--20664},
  year      = {2024}
}

@inproceedings{morgenstern2024compact,
  author    = {Wieland Morgenstern and Florian Barthel and Anna Hilsmann and Peter Eisert},
  title     = {Compact {3D} Scene Representation via Self-organizing {Gaussian} Grids},
  booktitle = ECCV,
  pages={18--34},
  year      = {2024}
}

@inproceedings{niedermayr2024compressed,
  author    = {Simon Niedermayr and Josef Stumpfegger and R{\"u}diger Westermann},
  title     = {Compressed {3D} {Gaussian} Splatting for Accelerated Novel View Synthesis},
  booktitle = CVPR,
  pages={10349--10358},
  year      = {2024}
}

@article{ren2025octreegs,
  author    = {Ren, Kerui and Jiang, Lihan and Lu, Tao and Yu, Mulin and Xu, Linning and Ni, Zhangkai and Dai, Bo},
  title     = {{Octree-GS}: Towards Consistent Real-time Rendering with {LOD}-structured {3D} {Gaussians}},
  journal   = PAMI,
  year      = {2025}
}

@inproceedings{ali2025elmgs,
  author    = {Ali, Muhammad Salman and Bae, Sung-Ho and Tartaglione, Enzo},
  title     = {{ElmGS}: Enhancing memory and computation scalability through compression for {3D} {Gaussian} splatting},
  booktitle = {Winter Conf. Appl. Comput. Vis.},
  pages     = {2591--2600},
  year      = {2025}
}

@inproceedings{zhang2018unreasonable,
  author    = {Zhang, Richard and Isola, Phillip and Efros, Alexei A and Shechtman, Eli and Wang, Oliver},
  title     = {The unreasonable effectiveness of deep features as a perceptual metric},
  booktitle = CVPR,
  pages     = {586--595},
  year      = {2018}
}

@article{chen2026feedforward,
  author  = {Chen, Yihang and Wu, Qianyi and Li, Mengyao and Lin, Weiyao and Hou, Junhui and Harandi, Mehrtash and Cai, Jianfei},
  title   = {Feedforward compression of static and streamable {3D} {Gaussian} splatting},
  journal = TCSVT,
  year    = {2026}
}

@inproceedings{chen2025megs,
  title={{MEGS²}: Memory-Efficient Gaussian Splatting via Spherical Gaussians and Unified Pruning},
  author={Chen, Jiarui and Chen, Yikeng and Zou, Yingshuang and Huang, Ye and Wang, Peng and Liu, Yuan and Sun, Yujing and Wang, Wenping},
  booktitle =ICLR,
  year={2026}
}

@inproceedings{du2026mobile,
  title={Mobile-GS: Real-time Gaussian Splatting for Mobile Devices},
  author={Du, Xiaobiao and Wang, Yida and Zhan, Kun and Yu, Xin},
  booktitle =ICLR,
  year={2026}
}

@inproceedings{bai2026plug,
  title={Plug-and-Play Optimization for 3D Gaussian Splatting Compression: Distribution Regularization, Probabilistic Pruning and Detail Compensation},
  author={Bai, Tian and Qiu, Zheng and Chen, Haojie and Dai, Ziyang},
  booktitle = AAAI,
  volume={40},
  number={4},
  pages={2372--2380},
  year={2026}
}

@article{kopanas2021point,
  author  = {Kopanas, Georgios and Philip, Julien and Leimk{\"u}hler, Thomas and Drettakis, George},
  title   = {Point-based neural rendering with per-view optimization},
  journal = CGF,
  volume  = {40},
  number  = {4},
  pages   = {29--43},
  year    = {2021}
}

@inproceedings{chen2025dashgaussian,
  title={Dashgaussian: Optimizing 3d gaussian splatting in 200 seconds},
  author={Chen, Youyu and Jiang, Junjun and Jiang, Kui and Tang, Xiao and Li, Zhihao and Liu, Xianming and Nie, Yinyu},
  booktitle = CVPR,
  pages={11146--11155},
  year={2025}
}

@article{ren2025fastgs,
  title={FastGS: Training 3D Gaussian Splatting in 100 Seconds},
  author={Ren, Shiwei and Wen, Tianci and Fang, Yongchun and Lu, Biao},
  booktitle = CVPR,
  pages={26094--26103},
  year={2026}
}

@inproceedings{wang2025steepest,
  title={Steepest descent density control for compact 3D Gaussian splatting},
  author={Wang, Peihao and Wang, Yuehao and Wang, Dilin and Mohan, Sreyas and Fan, Zhiwen and Wu, Lemeng and Cai, Ruisi and Yeh, Yu-Ying and Wang, Zhangyang and Liu, Qiang and others},
  booktitle= CVPR,
  pages={26663--26672},
  year={2025}
}

@article{chen2025hacplus,
  title   = {{HAC++}: Towards 100X Compression of 3D Gaussian Splatting},
  author    = {Yihang Chen and Qianyi Wu and Weiyao Lin and Mehrtash Harandi and Jianfei Cai},
  journal = PAMI,
  year    = {2025},
  volume  = {47},
  number  = {11},
  pages   = {10210--10226},
}

@inproceedings{rota2024revising,
  title={Revising densification in gaussian splatting},
  author={Rota Bul{\`o}, Samuel and Porzi, Lorenzo and Kontschieder, Peter},
  booktitle = ECCV,
  pages={347--362},
  year={2024}
}

@inproceedings{zhang2024pixel,
  title={Pixel-gs: Density control with pixel-aware gradient for 3d gaussian splatting},
  author={Zhang, Zheng and Hu, Wenbo and Lao, Yixing and He, Tong and Zhao, Hengshuang},
  booktitle = ECCV,
  pages={326--342},
  year={2024}
}

@inproceedings{ye2024absgs,
  title={Absgs: Recovering fine details in 3d gaussian splatting},
  author={Ye, Zongxin and Li, Wenyu and Liu, Sidun and Qiao, Peng and Dou, Yong},
  booktitle = ACMMM,
  pages={1053--1061},
  year={2024}
}

@inproceedings{cheng2024gaussianpro,
  title={Gaussianpro: 3d gaussian splatting with progressive propagation},
  author={Cheng, Kai and Long, Xiaoxiao and Yang, Kaizhi and Yao, Yao and Yin, Wei and Ma, Yuexin and Wang, Wenping and Chen, Xuejin},
  booktitle = ICML,
  pages={8123--8140},
  year={2024}
}

@inproceedings{yu2024mip,
  title={Mip-splatting: Alias-free 3d gaussian splatting},
  author={Yu, Zehao and Chen, Anpei and Huang, Binbin and Sattler, Torsten and Geiger, Andreas},
  booktitle = CVPR,
  pages={19447--19456},
  year={2024}
}

@inproceedings{huang20242d,
  title={2d gaussian splatting for geometrically accurate radiance fields},
  author={Huang, Binbin and Yu, Zehao and Chen, Anpei and Geiger, Andreas and Gao, Shenghua},
  booktitle={ACM SIGGRAPH 2024},
  pages={1--11},
  year={2024}
}

@article{kheradmand20243d,
  title={3d gaussian splatting as markov chain monte carlo},
  author={Kheradmand, Shakiba and Rebain, Daniel and Sharma, Gopal and Sun, Weiwei and Tseng, Yang-Che and Isack, Hossam and Kar, Abhishek and Tagliasacchi, Andrea and Yi, Kwang Moo},
  journal = NeurIPS,
  volume={37},
  pages={80965--80986},
  year={2024}
}

@article{gordon2012karush,
  title={Karush-kuhn-tucker conditions},
  author={Gordon, Geoff and Tibshirani, Ryan},
  journal={Optimization},
  volume={10},
  number={725/36},
  pages={725},
  year={2012}
}

@article{ormazabal1995law,
  title={The law of diminishing marginal utility in Alfred Marshall's principles of economics},
  author={Ormazabal, Kepa M},
  journal={Journal of the History of Economic Thought},
  volume={2},
  number={1},
  pages={91--126},
  year={1995},
  publisher={Taylor \& Francis}
}

@book{nocedal2006numerical,
  title={Numerical optimization},
  author={Nocedal, Jorge and Wright, Stephen J},
  publisher={Springer}
}

@inproceedings{xiangli2022bungeenerf,
  title={Bungeenerf: Progressive neural radiance field for extreme multi-scale scene rendering},
  author={Xiangli, Yuanbo and Xu, Linning and Pan, Xingang and Zhao, Nanxuan and Rao, Anyi and Theobalt, Christian and Dai, Bo and Lin, Dahua},
  booktitle = ECCV,
  pages={106--122},
  year={2022}
}

@article{wang2004image,
  title={Image quality assessment: from error visibility to structural similarity},
  author={Wang, Zhou and Bovik, Alan C and Sheikh, Hamid R and Simoncelli, Eero P},
  journal= TIP,
  volume={13},
  number={4},
  pages={600--612},
  year={2004},
  publisher={IEEE}
}

@article{chen2026refine,
  title={REFINE: Super-efficient 3D Gaussian Splatting Pruning via Rendering-Free Primitive Importance},
  author={Chen, Zhang and Wan, Shuai and Yu, Mengting and Yang, Fuzheng and Hou, Junhui},
  journal={arXiv preprint arXiv:2606.09074},
  year={2026}
}

@inproceedings{mallick2024taming,
  title={Taming 3dgs: High-quality radiance fields with limited resources},
  author={Mallick, Saswat Subhajyoti and Goel, Rahul and Kerbl, Bernhard and Steinberger, Markus and Carrasco, Francisco Vicente and De La Torre, Fernando},
  booktitle = SIGGRAPH ,
  pages={1--11},
  year={2024}
}

@inproceedings{yang2026rap,
  title={RAP: Fast Feedforward Rendering-Free Attribute-Guided Primitive Importance Score Prediction for Efficient 3D Gaussian Splatting Processing},
  author={Yang, Kaifa and Yang, Qi and Xu, Yiling and Li, Zhu},
  booktitle = CVPR,
  pages={33323--33332},
  year={2026}
}

@article{fang2026efficient,
  author  = {Guangchi Fang and Bing Wang},
  title   = {Efficient Scene Modeling via Structure-Aware and Region-Prioritized 3D Gaussians},
  journal = PAMI,
  year    = {2026},
  volume  = {48},
  number  = {4},
  pages   = {4623--4641},
}

@article{wang2026freesplatplus,
  author  = {Yunsong Wang and Tianxin Huang and Hanlin Chen and Gim Hee Lee},
  title   = {{FreeSplat++}: Generalizable 3D Gaussian Splatting for Efficient Indoor Scene Reconstruction},
  journal = PAMI,
  year    = {2026},
  volume  = {48},
  number  = {7},
  pages   = {7749--7765},
}

@article{shen2025gamba,
  author  = {Qiuhong Shen and Zike Wu and Xuanyu Yi and Pan Zhou and Hanwang Zhang and Shuicheng Yan and Xinchao Wang},
  title   = {Gamba: Marry Gaussian Splatting with Mamba for Single-View 3D Reconstruction},
  journal = PAMI,
  year    = {2025},
}

@article{zhou2025gpsgaussianplus,
  author  = {Boyao Zhou and Shunyuan Zheng and Hanzhang Tu and Ruizhi Shao and Boning Liu and Shengping Zhang and Liqiang Nie and Yebin Liu},
  title   = {{GPS-Gaussian+}: Generalizable Pixel-Wise 3D Gaussian Splatting for Real-Time Human-Scene Rendering from Sparse Views},
  journal = PAMI,
  year    = {2025},
}

@article{wu2025deferredgs,
  author  = {Tong Wu and Jia-Mu Sun and Yu-Kun Lai and Yuewen Ma and Leif Kobbelt and Lin Gao},
  title   = {{DeferredGS}: Decoupled and Relightable Gaussian Splatting with Deferred Shading},
  journal = PAMI,
  year    = {2025},
  volume  = {47},
  number  = {8},
  pages   = {6307--6319},
}

@article{lei2025gaussnav,
  author  = {Xiaohan Lei and Min Wang and Wengang Zhou and Houqiang Li},
  title   = {{GaussNav}: Gaussian Splatting for Visual Navigation},
  journal = PAMI,
  year    = {2025},
  volume  = {47},
  number  = {5},
  pages   = {4108--4121},
}
}

\begin{IEEEbiography}[{\includegraphics[width=1in,height=1.25in,clip,keepaspectratio]{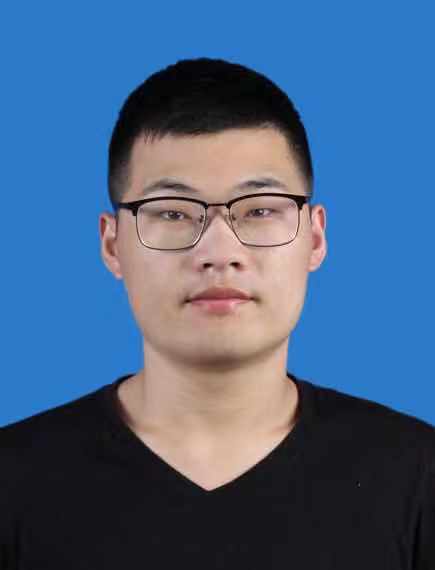}}]{Zhang Chen} 
received the B.E. degree in Electronics and Information Engineering and the M.E. degree in Signal and Information Processing from Northwestern Polytechnical University, Xi’an, China, in 2019 and 2022, respectively. He is currently pursuing the Ph.D. degree in Information and Communication Engineering at the same university. Since 2026, he has been a Research Assistant with the Department of Computer Science, City University of Hong Kong, Hong Kong SAR, China. His current research interests include 3DGS pruning, 3DGS optimization, multimedia quality assessment, and 3D reconstruction.
\end{IEEEbiography}

\begin{IEEEbiography}
[{\includegraphics[width=1in,height=1.2in,clip,keepaspectratio]{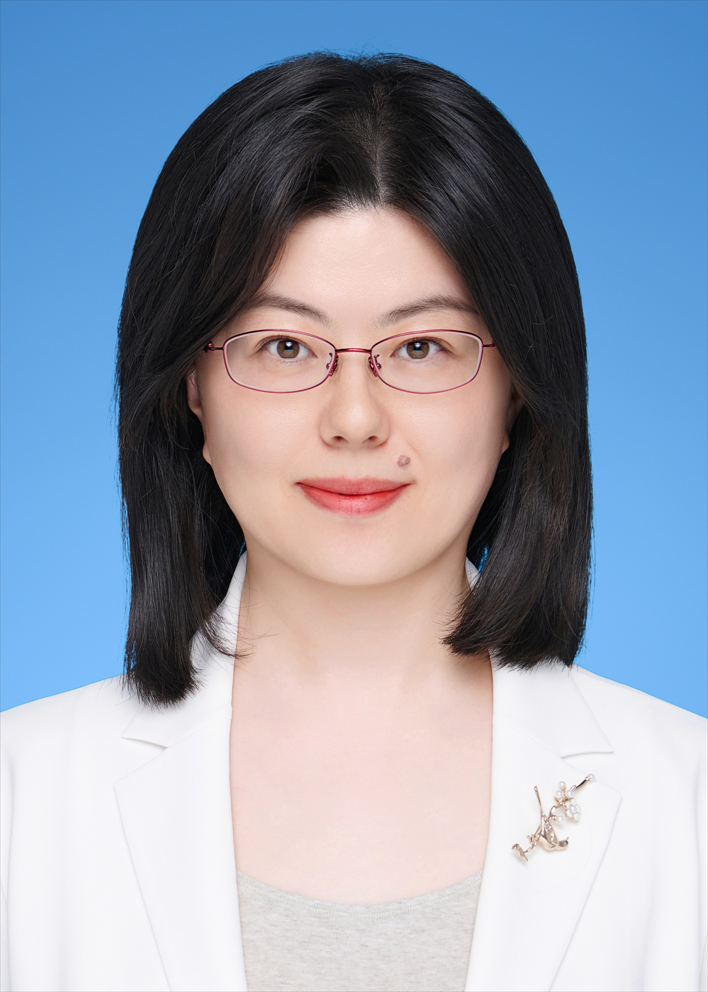}}]{Shuai Wan} (Member, IEEE) received the B.E. degree in Telecommunication Engineering and the M.E. degree in Communication and Information System from Xidian University, Xi’an, China, in 2001 and 2004, respectively, and obtained the Ph.D. in Electronic Engineering from Queen Mary, University of London in 2007. She is currently a Professor at Northwestern Polytechnical University in Xi’an, China. Previously, she served as a Professor at the Royal Melbourne Institute of Technology in Melbourne, Australia, from 2016 to 2025. Her research interests include scalable/multiview video coding, video quality assessment and hyperspectral image compression.
\end{IEEEbiography}

\begin{IEEEbiography}
[{\includegraphics[width=1in,height=1.2in,clip,keepaspectratio]{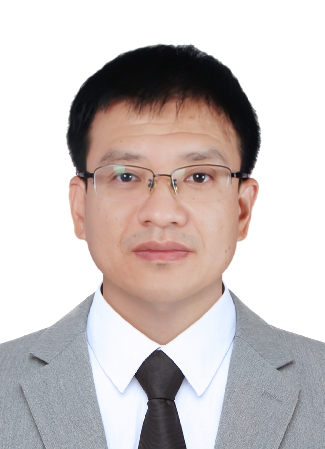}}]{Fuzheng Yang}(Member, IEEE) received the B.E. degree in Telecommunication Engineering, the M.E. degree and the Ph.D. in Communication and Information System from Xidian University, Xi’an, China, in 2000, 2003 and 2005, respectively. He became a lecturer and an Associate Professor in Xidian University in 2005 and 2006, respectively. He has been a professor of communications engineering with Xidian University since 2012. He is also an Adjunct Professor of School of Engineering in RMIT University. During 2006-2007, he served as a visiting scholar and postdoctoral researcher in Department of Electronic Engineering in Queen Mary, University of London. His research interests include video quality assessment, video coding and multimedia communication.
\end{IEEEbiography}

\begin{IEEEbiography}
[{\includegraphics[width=1in,height=1.2in,clip,keepaspectratio]{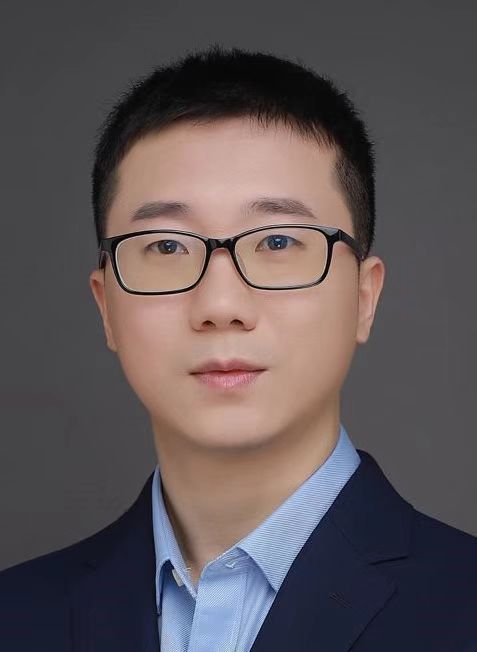}}]{Jiazhi Xia} is a Professor in the School of Computer Science and Engineering, Central South University.
His research interests include visualization, geometry processing, and 3D vision.
\end{IEEEbiography}

\begin{IEEEbiography}
[{\includegraphics[width=1in,height=1.2in,clip,keepaspectratio]{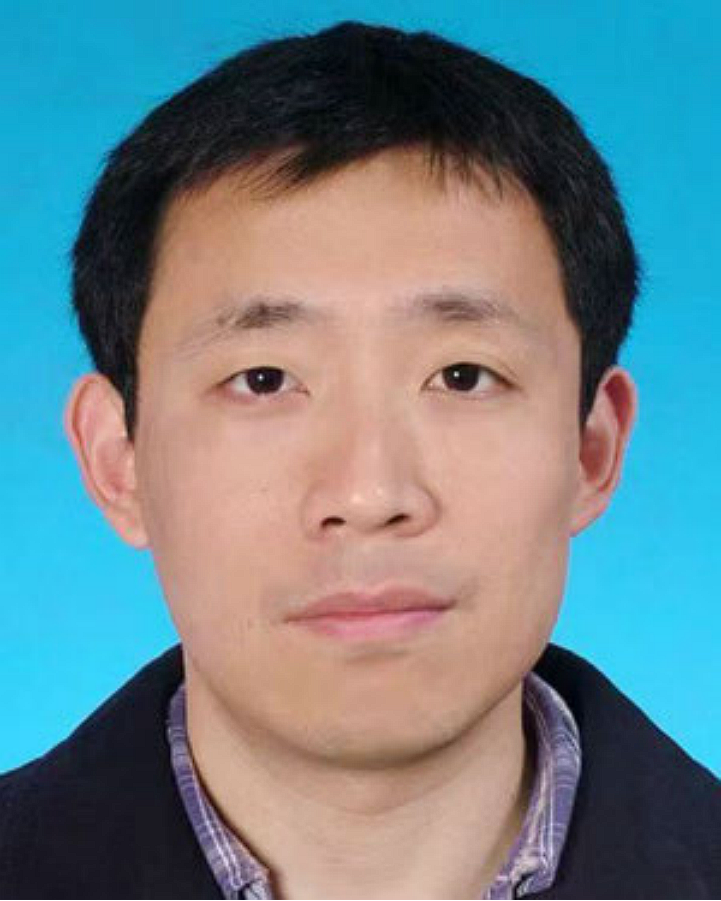}}] {Weiyao Lin}(Senior Member, IEEE) received the B.E. and M.E. degrees from Shanghai Jiao Tong University, Shanghai, China, in 2003 and 2005, respectively, and the Ph.D. degree from the University of Washington, Seattle, WA, USA, in 2010, all in electrical engineering. He is currently a Professor with the Department of Electronic Engineering, Shanghai Jiao Tong University. He has authored or co-authored more than 100 technical papers in top journals/conferences, including IEEE TRANSACTIONS ON PATTERN ANALYSIS AND MACHINE
INTELLIGENCE, International Journal of Computer Vision, IEEE TRANSACTIONS ON IMAGE PROCESSING, CVPR, and ICCV. He holds more than 20 patents. His research interests include video/image analysis, computer vision, and video/image processing applications.
\end{IEEEbiography}

\begin{IEEEbiography}
[{\includegraphics[width=0.95in,height=1.15in,clip,keepaspectratio]{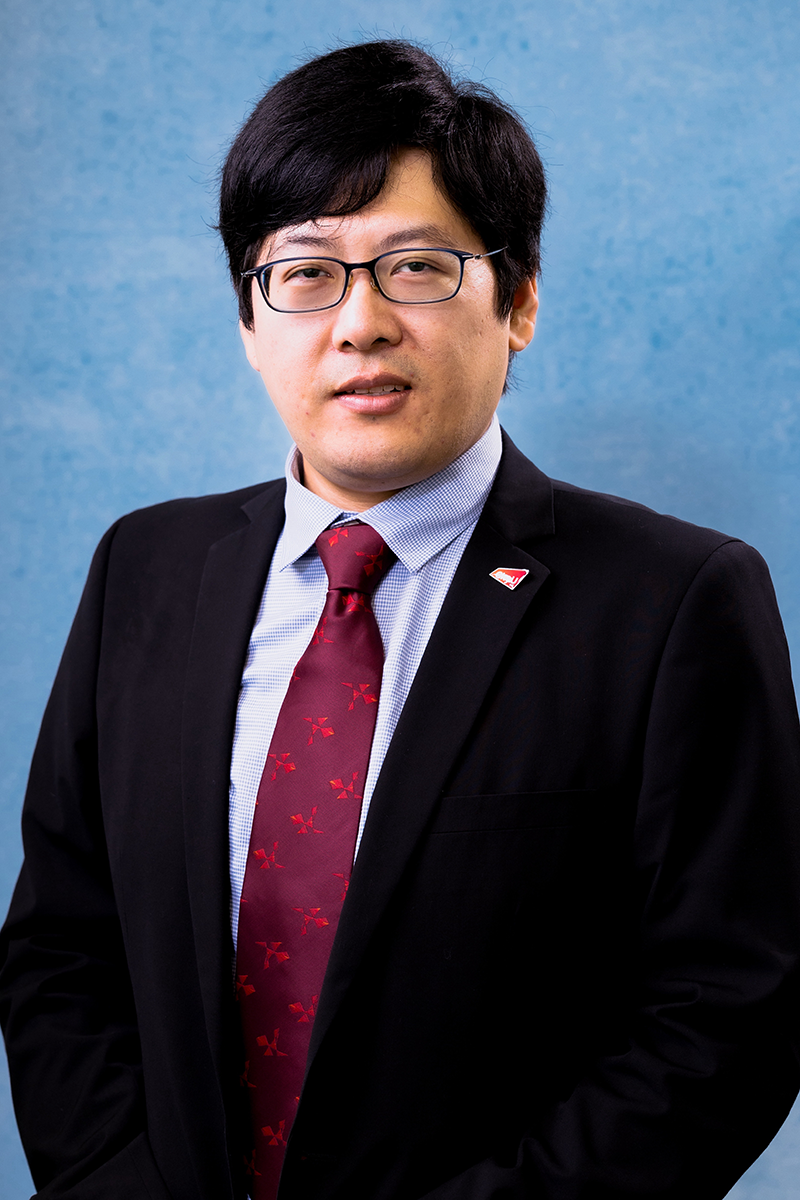}}]{Junhui Hou}(Senior Member, IEEE) is a Professor with the Department of Computer Science, City University of Hong Kong His research interests include multidimensional visual computing, such as light field, hyperspectral, geometry, and event data. He received the Early Career Award from the Hong Kong Research Grants Council in 2018, IEEE Multimedia Rising Star Award in 2023, the Excellent Young Scientists Fund from NSFC in 2024, and the IEEE TIP Best Paper Award in 2025. He is serving as a Senior Area Editor for IEEE TIP and an Associate Editor for IEEE TVCG and TMM. He served as an Associate Editor for IEEE TIP and TCSVT.
\end{IEEEbiography}

\end{document}